\documentclass{article}

\usepackage{dharma_ai}
\tcbuselibrary{breakable}

\setcompanylogo{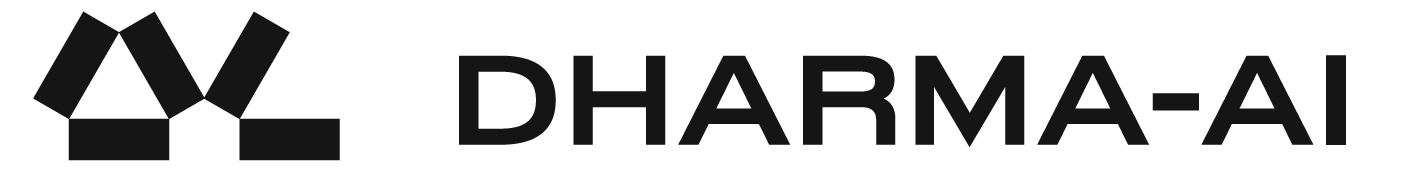}
\setcompanylogoheight{16pt}

\usepackage[utf8]{inputenc}
\usepackage[T1]{fontenc}

\usepackage{hyperref}
\usepackage{url}
\usepackage{fontawesome5}
\usepackage{booktabs}
\usepackage{amsfonts}
\usepackage{amsmath}
\usepackage{nicefrac}
\usepackage{pdflscape}
\usepackage{microtype}
\usepackage{graphicx}
\graphicspath{{.}}
\usepackage{xcolor}
\usepackage{siunitx}

\newcommand{\hficon}{\IfFileExists{huggingface_logo.png}{\raisebox{-0.15em}{\includegraphics[height=1.2em]{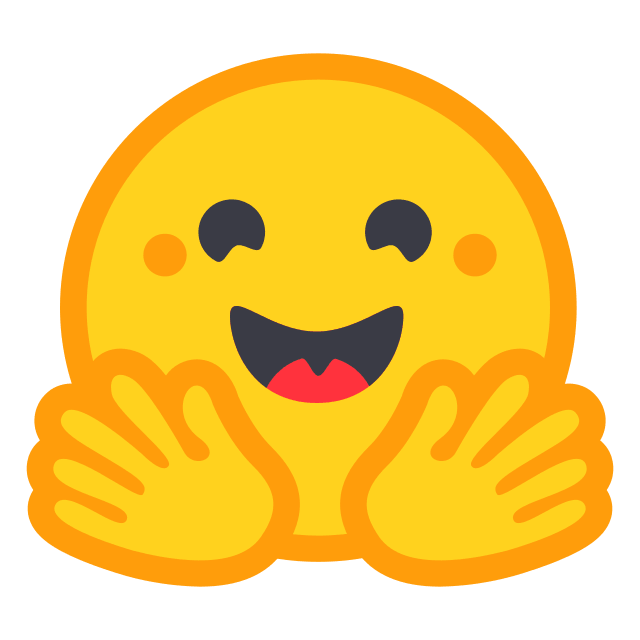}}}{HF}}
\newcommand{\dharmaicon}{\IfFileExists{logo_solo_dharmaai.png}{\raisebox{-0.40em}{\includegraphics[height=1.5em]{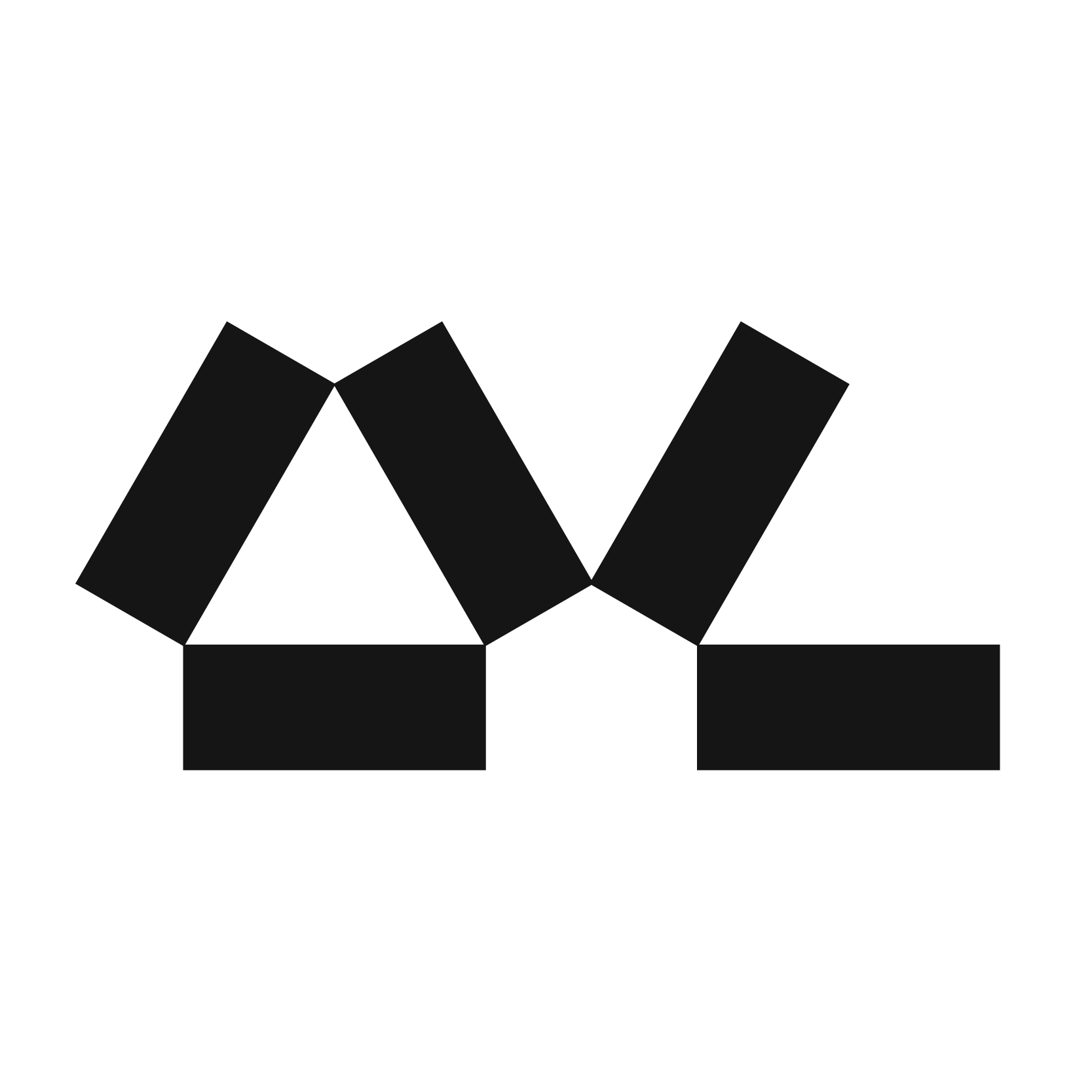}}}{Dharma}}
\usepackage{tabularx}
\usepackage{multirow}
\usepackage[table]{xcolor}
\usepackage{tikz}
\usetikzlibrary{positioning, shapes.geometric, arrows.meta, shadows}

\usepackage{longtable}
\usepackage{array}
\usepackage{afterpage}
\usepackage{capt-of}

\tikzset{
    base/.style = {draw, font=\small, align=center, minimum height=1cm, transition},
    data/.style = {base, cylinder, shape border rotate=90, fill=blue!10, minimum width=2cm},
    process/.style = {base, rectangle, rounded corners, fill=gray!10, text width=2.5cm},
    core/.style = {base, rectangle, fill=orange!15, text width=2.5cm, draw=orange, line width=1pt},
    result/.style = {base, ellipse, fill=green!10, text width=2cm, draw=green!50!black},
    arrow/.style = {->, >=Stealth, thick, line width=0.8pt}
}

\usepackage{listingsutf8}
\usepackage{pgfplots}
\pgfplotsset{compat=1.18}

\AtBeginDocument{%

  \renewcommand{\today}{%
    \ifcase\month\or January\or February\or March\or April\or May\or June\or July\or August\or September\or October\or November\or December\fi
    \space\number\day, \number\year%
  }%
}

\begin{document}

\title{DharmaOCR: Specialized Small Language Models for Structured OCR that outperform Open-Source and Commercial Baselines}

\author{
  Gabriel Pimenta de Freitas Cardoso \\
  Dharma-AI \\
  \texttt{gabriel.pimenta@dharma-ai.com} \\
  \And
  Caio Lucas da Silva Chacon \\
  Dharma-AI \\
  \texttt{caio.chacon@dharma-ai.com} \\
  \And
  Jonas Felipe da Fonseca Oliveira \\
  Dharma-AI \\
  \texttt{jonas.oliveira@dharma-ai.com} \\
  \And
  Paulo Henrique de Medeiros Araujo \\
  Dharma-AI \\
  \texttt{paulo.araujo@dharma-ai.com}
}

\maketitle

\begin{abstract}
This manuscript introduces DharmaOCR Full and Lite, a pair of specialized small language models (SSLMs) for structured OCR that jointly optimize transcription quality, generation stability, and inference cost. It also presents \textbf{DharmaOCR-Benchmark}, a benchmark that covers printed, handwritten, and legal/administrative documents, and proposes a unified evaluation protocol that measures fidelity and structure while explicitly tracking text degeneration as a first-class benchmark metric (alongside unit cost). Beyond reporting degeneration rates, the manuscript empirically shows degeneration is not merely a quality failure, since it materially worsens production performance by increasing response time, reducing throughput, and inflating computational cost due to abnormally long generations. To the best of the author's knowledge, as a methodological contribution, this is the first application of Direct Preference Optimization (DPO) for OCR, explicitly using degenerate generations as \textit{rejected} examples to penalize looping behavior. Combined with Supervised Fine-Tuning (SFT) for enforcing a strict JSON schema (header, margin, footer, and text), DPO consistently reduces degeneration rate across model families (up to 87.6\% relative) while preserving or improving extraction quality. The resulting models, namely, \textbf{DharmaOCR Full} (7B) and \textbf{DharmaOCR Lite} (3B), set a new state-of-the-art on DharmaOCR-Benchmark, outperforming each open-source and commercial baseline model evaluated regarding extraction quality, reaching 0.925 and 0.911 scores with 0.40\% and 0.20\% degeneration rates. AWQ quantization reduced up to 22\% per-page cost with negligible quality loss, enabling a strong quality-cost trade-off in comparison to proprietary OCR APIs and open-source alternatives.
\end{abstract}

\begin{center}
\small
\begin{tabular}{@{}ll@{}}
\hficon\ Benchmark: & \href{https://huggingface.co/datasets/Dharma-AI/DharmaOCR-Benchmark}{Dharma-AI/DharmaOCR-Benchmark}\\
\hficon\ Model: & \href{https://huggingface.co/Dharma-AI/DharmaOCR-Lite}{Dharma-AI/DharmaOCR-Lite}\\
\dharmaicon\ Website: & \href{https://dharma-ai.com.br/}{dharma-ai.com.br}\\
\end{tabular}
\end{center}

\section{Introduction}

Optical character recognition (OCR) plays a central role in document processing systems by enabling conversion of illegible or partially legible documents into structured textual representations amenable to computational processing. It was essential for downstream tasks applied to the content of this document, such as information retrieval, entity extraction, classification, summarization, semantic search, and other Natural Language Processing (NLP) applications \cite{smith2007tesseract,shi2016crnn}.

Beyond being a pre-processing step, high-quality OCR is a key enabler for deploying Generative AI on real-world organizational knowledge. A substantial share of valuable information remains locked in unstructured artifacts (e.g., paper archives, scanned PDFs, images, and legacy files) and must be converted into a machine-readable text prior to being reliably indexed, retrieved, or used for training and evaluation of models. In this context, the well-known “garbage in, garbage out” principle still applies, since errors introduced during text digitization propagate to downstream generative systems, degrading factuality, controllability, and overall output quality \cite{sculley2015hiddentechdebt}.

Classic OCR systems were historically developed as modular pipelines comprising text detection, segmentation, and character recognition. Architectures based on Convolutional Recurrent Neural Networks (CRNNs), trained with Connectionist Temporal Classification (CTC) became widely adopted for enabling end-to-end training \cite{graves2006ctc,shi2016crnn}. However, those systems left relevant performance gaps when faced with heterogeneous documents, different document domains, and structural variations frequently encountered in real-world applications \cite{greif2025multimodalllmsocrocr, xu2020layoutlm,xu2021layoutlmv2}.

In recent years, multimodal Large Language Model (LLM)-based document information extraction architectures, such as GPT-4o and Gemini 2.5 Pro, have been explored through the integration of deep visual perception and broad linguistic knowledge across domains for enhancing the understanding of complex and diverse layouts \cite{xu2021layoutlmv2,kim2022donut}.  Those models improve the ability to handle varied structures; however, they introduce new practical challenges primarily related to high computational costs. Another limitation is LLMs are in general generalist, which may lead to a lack of specialization for specific domains.

As an alternative, Small Language Models (SLMs) have emerged as deep models substantially smaller than LLMs in terms of parameters, with architectures of up to approximately 15 billion parameters (despite no universal definition) \cite{subramanian2025smalllanguagemodelsslms}. Such parameter count is at least two orders of magnitude smaller than that of the largest LLMs. Although those models offer shorter per-instance inference time and lower cost per request \cite{zhou2023slm}, they may lose performance in tasks due to their reduced computational capacity.

In contrast, Specialized SLMs (SSLMs) are small language models intentionally optimized for a specific task and/or domain, rather than trained to be broadly capable general-purpose assistants. Whereas generalist LLMs and generic SLMs ar trained to absorb and follow a wide range of instructions across heterogeneous domains, SSLMs are trained so that their limited parameters and computational resources are concentrated on a single (or a small set of) target problems \cite{pecher2026comparingsslm}.

By allocating capacity to target distribution, SSLMs can outperform much larger LLMs in the intended application while matching or undercutting inference time and cost of generic SLMs, thus remaining far cheaper than LLMs \cite{pecher2026comparingsslm}. For specific high-volume applications, such as OCR, SSLMs provide a practical alternative that increases task performance while substantially reducing inference latency and operational cost.

This manuscript addresses the development of DharmaOCR Full and DharmaOCR Lite, two state-of-the-art SSLMs for OCR that outperformed all open-source and commercial models evaluated in the proposed benchmark. They result from extensive experiments combining different model families (e.g., Qwen and Gemma) at multiple parameter scales and specializing them through combinations of fine-tuning techniques, including Supervised Fine-Tuning (SFT) and Direct Preference Optimization (DPO). According to the results, in addition to achieving the best benchmark performance, the models also offer a reduced per-page unit cost when compared to similar models.

\subsection{Contributions}
\begin{itemize}
    \item Release of DharmaOCR-Benchmark, a dataset benchmark for evaluations of OCR models in contexts underexplored in other benchmarks;
    \item Introduction of DharmaOCR Full and Lite, state-of-the-art OCR models that outperformed the main open-source and commercial baselines evaluated in the proposed benchmark;
    \item An in-depth discussion of text degeneration, assessing its impact on performance and cost in real AI systems and showing the effectiveness of the specialization techniques adopted to mitigate that issue;
    \item Introduction of a DPO fine-tuning strategy explicitly targeting reduction of text degeneration and empirical evidence of substantial gains;
    \item Proposal of use of SFT + DPO for fine-tuning OCR models for the first time, according to the authors’ knowledge, showing improvements in quality and cost and a reduction in text degeneration;
    \item Presentation of results from an extensive suite of experiments with models of varying architectures and sizes with different specialization techniques, compared in a systematic and unified way;
    \item Demonstration specializing OCR models for narrower domains starting from OCR-specialized models achieve better results than starting from vanilla models of same architecture.
\end{itemize}

\begin{figure}[htbp]
    \centering
    \includegraphics[width=0.97\linewidth]{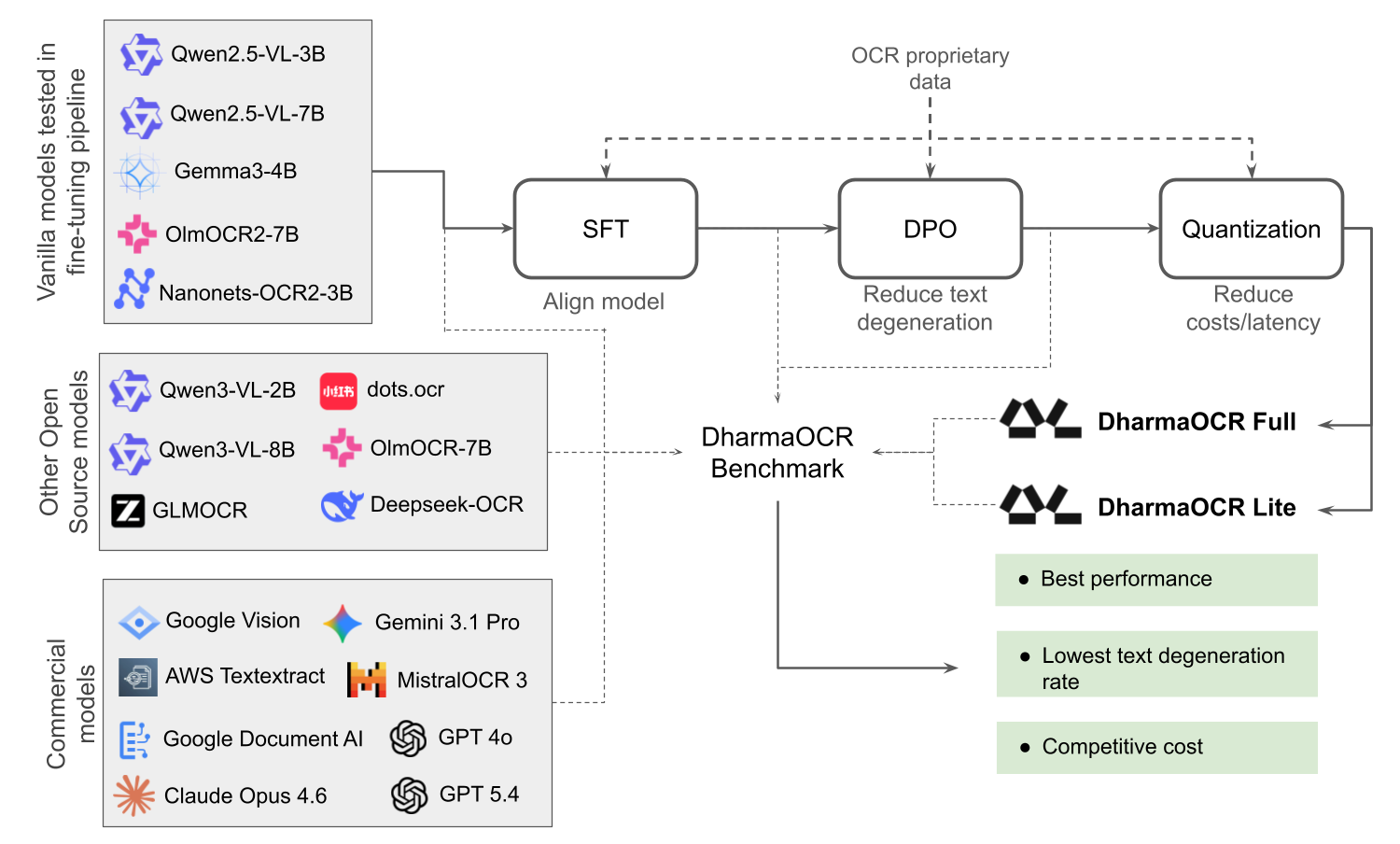}
    \caption{Synthesis of the proposed approach, key contributions, and results, illustrating the progression from vanilla models to improvements enabled by specialization (SFT+DPO) and text degeneration mitigation, outperforming other models on DharmaOCR-Benchmark.}
    \label{fig:init_image}
\end{figure}

\section{Related Work}

Some SSLMs have recently been proposed for OCR for transcribing page text while preserving its structure (reading order, tables, lists, equations) into formats such as linearized text - an example is olmOCR \cite{poznanski2025olmocr}, a model obtained by fine-tuning Qwen2-VL-7B-Instruct \cite{wang2024qwen2vlenhancingvisionlanguagemodels} on data from books and articles with diverse layouts, including tables and equations. In a subsequent iteration, called olmOCR 2 \cite{poznanski2025olmocr2unittest}, the model is improved by both Qwen2.5-VL-7B-Instruct \cite{bai2025qwen25vltechnicalreport} as the base architecture and addition of an extra fine-tuning step, namely, OCR unit tests for fine-tuning via Reinforcement Learning with Verifiable Rewards (RLVR).

In parallel, structured-output approaches have gained traction with models such as Nanonets-OCR2-3B \cite{nanonets2025ocr2}, image-to-text models that convert documents directly into Markdown. Additionally, OCR-specific encoder--decoder architectures, including DeepSeek-OCR \cite{wei2025deepseekocrcontextsopticalcompression}, DeepSeek-OCR-2 \cite{wei2026deepseekocr2visualcausal}, and GLM-OCR \cite{glm_ocr_hf} have emerged, differing from the classic LLM/SLM architectures that are typically decoder-only \cite{zhang2025encoderdecoderdecoderonlyrevisitingencoderdecoder}.

All aforementioned models have shown strong potential for extracting text from images and PDFs, outperforming, on several benchmarks, traditional commercial systems and closed-source LLMs such as GPT-4o and Gemini 2.5 Pro. However, they still leave relevant gaps.

One of these gaps is the limited ability to cover different domains. In general, smaller models tend to show poorer out-of-distribution (OoD) generalization and lower sample efficiency than larger ones \cite{kaplan2020scalinglaws}. An SLM fine-tuned for OCR can outperform general-purpose LLMs in domains covered by a fine-tuning dataset, as reflected in the benchmarks reported in the corresponding papers. However, such performance can degrade when the application domain shifts (e.g. when those models are applied to languages underrepresented in the training data).

Another gap refers to text degeneration, a known phenomenon in autoregressive language models that is still present in modern SLMs and LLMs \cite{holtzman2020curiouscaseneuraltext, kwon2026llmsthinkdonttell}, but is not adequately discussed in papers that propose the aforementioned models. The phenomenon can lead to nonsensical outputs, often involving loops with repeated tokens or sentences indefinitely until the model hits the configured token generation limit \cite{welleck2019neural, Yao_2025}, sharply reducing the reliability of OCR models and degrading system performance in terms of latency, throughput, and cost.

This manuscript addresses both gaps. Multiple specialization paths were evaluated for open-source SLMs towards their adaptation to undercovered domains, showing substantial improvements on a benchmark created for those domains. In addiction, a drastic reduction in text degeneration at the end of the proposed fine-tuning pipeline was demonstrated, including the use of DPO tailored to this, a novel strategy for OCR models (to the best of the authors' knowledge) was introduced.

\section{Text Degeneration}
\label{sec:degeneracao-textual}

Text degeneration in language models refers to the tendency of a model to produce repetitive or generic sequences, or to get trapped in local high-probability loops \cite{holtzman2020curiouscaseneuraltext, welleck2019neural, Yao_2025, lee-etal-2023-local, li2023repetitionrepetitionoutunderstanding, fu2021theoreticalanalysisrepetitionproblem, wang2024mitigatinglanguagemismatchrepetition, chiang2021relatingneuraltextdegeneration}. The phenomenon does not have a single root cause with a direct fix; therefore, several studies have been conducted to mitigating it. Some have have reduced text degeneration by proposing new decoding strategies that help LLMs escape such loops driven by self-reinforcement and high marginal probability \cite{holtzman2020curiouscaseneuraltext, lee-etal-2023-local}. Other approaches have argued repetitive patterns are inherent to language itself, which yields repetitive patterns in the training data and increases repetitions in model generations, suggesting degeneration also has a structural component associated with the empirical data distribution \cite{welleck2019neural, li2023repetitionrepetitionoutunderstanding, fu2021theoreticalanalysisrepetitionproblem}. Other studies have aimed to understand the internal mechanisms in autoregressive neural networks that give rise to that phenomenon during inference \cite{Yao_2025, wang2024mitigatinglanguagemismatchrepetition, chiang2021relatingneuraltextdegeneration}.

Degenerate scenarios show a positive-feedback mechanism (self-reinforcement), in which the conditional probability of a previously generated token increases progressively as it is repeated. Holtzman et al. \cite{holtzman2020curiouscaseneuraltext} illustrated the probability of a repeated phrase growing with each new occurrence, creating a stable cycle of high likelihood. Formally, $P(x_{t+1}=w \mid x_{<t+1})$ increases if $x_t = w$, producing a self-reinforcing local probabilistic attractor \cite{welleck2019neural, Yao_2025}. The effect can be amplified by decoding and training strategies that optimize maximum likelihood, as well as by internal mechanisms that capture repetitive patterns \cite{li2023repetitionrepetitionoutunderstanding, wang2024mitigatinglanguagemismatchrepetition}. Figure~\ref{fig:degeneracao-textual-exemplo} displays a visual example of token- and sequence-level degeneration. Additional real examples can be found in Appendix \ref{apx:examples-degeneration}.

\begin{figure}[h]
  \centering
  \includegraphics[width=0.6\linewidth]{example_text_degeneration.png}
  \caption{Pictorial example of token- and sequence-level text degeneration, in which a single token (or token sequence) dominates the conditional distribution, producing repetitions indefinitely.}
  \label{fig:degeneracao-textual-exemplo}
\end{figure}

Therefore, degeneration should not be interpreted solely as a decoding failure. Instead, it appears to emerge from the interaction among multiple factors, including maximum-likelihood objective used during training and inference, internal geometry of activations that encode repetitive patterns from the training data, and decoding strategies that exploit high-probability regions of the distribution that are artificially created by network self-reinforcement.

\subsection{Effect}

Text degeneration is not only an output-quality issue, since it can often be handled with additional mechanisms in production systems (e.g., automatic retries that trigger different models or decoding strategies, streaming repetition detection, and similar safeguards). The core issue is system performance itself, mainly cost, response latency, and compute-resource usage. Those performance problems arise from two main factors.

First, all aforementioned mitigation mechanisms for handling degeneration in production introduce computational overhead due to both rerunning models and real-time repetition detection.

Second, even ignoring such safeguards, degenerate requests with long responses reduce the efficiency of the system that serves the model. In those situations, the model repeats a single token (or a small set of tokens) until it is stopped by an operational generation limit, since it does not emit the end-of-sequence (EOS) token in that loop. As a result, the output length of those requests becomes much larger than that of typical requests, demanding many more decoding steps, hence, leading to two negative effects on the system. The first is a request's response time is strongly correlated with its token count \cite{fu2024breaksequentialdependencyllm} - a single degenerate case can occupy enough time in the model's execution queue to answer multiple medium-length requests \cite{patke2025queuemanagementsloorientedlarge}.

The second effect is even when inference resource-optimization strategies (e.g., vLLM's PagedAttention \cite{vllm2026}) are used, memory consumption grows approximately linearly with total number of tokens processed per sequence \cite{kwon2023efficientmemorymanagementlarge}. Therefore, when degeneration occurs, it occupies a significant fraction of VRAM for a prolonged period. Since available memory is the main limiting factor for effective parallelism, it reduces the number of sequences that can be executed simultaneously in the dynamic batch, decreasing parallelization, hence, the system's aggregate throughput \cite{kwon2023efficientmemorymanagementlarge, agrawal2024tamingthroughputlatencytradeoffllm}. The impact is not restricted to the degenerate request, all requests executed in parallel are affected by that throughput reduction, making the system less efficient.

Those effects are demonstrated empirically in Appendix \ref{apx:impacto-degeneracao}, which shows a degenerate request can, in some cases, nearly double GPU time and cost required to answer a set of requests received by the system, even when retry mechanisms are ignored.

Therefore, the number of text degeneration events must be reduced in real production AI systems. Accordingly, studies that propose models and benchmarks should actively track that metric so as to quantify the impact of the phenomenon on user experience, response time, cost of systems that serve LLMs/SLMs, and the overall system reliability \cite{kwon2026llmsthinkdonttell}.

\section{Methodology for developing DharmaOCR}

The development of the models followed these steps: (i) selection of candidate models for fine-tuning and subsequent comparison; (ii) acquisition and structuring of training and evaluation data; (iii) training of selected models and concomitant monitoring of performance and convergence metrics; (iv) quantization of most promising models; and (v) evaluation and comparison of the developed models against competing models under a unified inference and metric protocol.

\subsection{Selected models}

The selection of models for fine-tuning aimed to balance two main axes, namely, general-purpose vanilla SLMs with no specialization and SLMs specialized for OCR with established performance. The list of selected models made enabled investigations of both incremental gains on top of domain-oriented architectures and potential of adaptation of models originally trained for broader tasks.

Among the considered models, those with competitive performance reported in the literature and in public benchmarks widely used by the community and with practical evidence of robustness in document scenarios were prioritized \cite{ocrbench_leaderboard}. Models that differed in architecture and parameter count, namely, olmOCR-2-7B \cite{poznanski2025olmocr2unittest}; Nanonets-OCR2-3B \cite{nanonets2025ocr2}; Qwen2.5-VL-7B-Instruct \cite{qwen2p5_vl_7b}; Qwen2.5-VL-3B-Instruct \cite{qwen2p5_vl_3b}; and gemma-3-4b-it \cite{gemma_3_4b_it} were also selected.

The first two are SLMs already specialized for OCR, whereas the other three are general-purpose multimodal models widely used across tasks, spanning different model families and parameter scales. olmOCR-2-7B and Nanonets-OCR2-3B were obtained by fine-tuning models from the Qwen2.5-VL family, respectively 7B-Instruct \cite{poznanski2025olmocr2unittest} and 3B-Instruct \cite{nanonets2025ocr2} variants. Therefore, Qwen2.5-VL was also included as a baseline, enabling a direct assessment of the impact of fine-tuning specialization. Moreover, that family is competitive on OCR-related tasks, even outperforming its successor Qwen3-VL on benchmarks \cite{qwen3-vl-2025, ocrbench_leaderboard}. Altogether, the set enables comparisons between specialization starting from a vanilla model and specialization starting from models previously tuned for a similar domain.

\subsection{Datasets}

The training dataset was built to provide typological diversity of pages for OCR, including a variety of layouts (e.g., plain-text pages, double-column pages, images, tables, forms, and other characteristics), different typographic styles (handwritten, typed, and mixed text), varying levels of visual noise (good, medium, and poor image quality), and a range of topics in the extracted text.

The corpus combines data from public sources such as arXiv and proprietary data generated internally, totaling $39\,680$ pages with text predominantly in Brazilian Portuguese. All sources underwent anonymization and sensitive-information removal procedures prior to incorporated into the dataset.

The corpus was partitioned into a training split and a held-out evaluation split. Multiple model families and training variants (FFT/LoRA; SFT/SFT+DPO) were trained and the final released models were those with best performance on the held-out evaluation split. The benchmark introduced in Section~\ref{sec:metodologia-avaliacao} is a separate dataset and was not used in any prior step, i.e., training, model selection, DPO pair construction, or quantization calibration.

A strategy that uses large LLMs for text extraction was adopted and followed by sampled human verification and validation for quality control for the labeling of the dataset. Three models, namely, \texttt{Claude Sonnet 4} \cite{anthropic2025claudesonnet4}, \texttt{llama-4-maverick-17b-128e-instruct-maas} \cite{meta2025llama4maverick}, and \texttt{gemini-2.5-pro} \cite{google2025gemini25pro} were used in that step.

The generated labels follow a strict JSON format, with explicit fields for \texttt{header}, \texttt{text}, \texttt{footer}, and \texttt{margin} containing the text of each field, as shown in Appendix \ref{apx:output-format}. The main reasons for the choice were:

\begin{itemize}
  \item \textbf{Verifiable field separation:} a JSON schema separated the main page fields in a verifiable way;
  \item \textbf{Focus on the main information:} inspired in \cite{poznanski2025olmocr} and \cite{poznanski2025olmocr2unittest}, in which the authors teach the model to extract the main page text and ignore low-information fields such as headers and footers, data were annotated in JSON with structured fields to enable the same separation;
  \item \textbf{Extraction completeness:} a limitation of the approach in \cite{poznanski2025olmocr} and \cite{poznanski2025olmocr2unittest} is, depending on the application context, important information in auxiliary fields may be ignored. Therefore, the above structure is proposed so that users can decide on whether to use the information in the header, footer, or margins according to their context.
\end{itemize}

\subsection{Supervised Fine-Tuning}
\label{subsec:SFT}

Supervised Fine-Tuning (SFT) was applied as the initial specialization method, towards bringing the model distribution closer to the document domain by maximizing likelihood between model outputs and training labels. Formally, the training objective minimizes Cross-Entropy Loss, or Negative Log-Likelihood (NLL), over the set of supervision pairs \(\{(x^{(i)}, y^{(i)})\}\):

\[
\mathcal{L}_{\text{NLL}} = -\sum_{i}\sum_{t=1}^{T^{(i)}} \log P_{\theta}\big(y^{(i)}_{t} \mid y^{(i)}_{<t}, x^{(i)}\big),
\]

where \(x^{(i)}\) denotes input and \(y^{(i)}\) is the reference output. The formulation is standard in supervised fine-tuning work for language models \cite{radford2018improving}.

Full fine-tuning (FFT) and parameter-efficient fine-tuning (PEFT), specifically LoRA \cite{hu2021lora}, were performed for each model for comparison of both approaches for the target specialization.

The main hyperparameters were defined via controlled experimentation. Learning rates Were evaluated in the $1\times10^{-4}$ to $1\times10^{-7}$ range and trained for two to five epochs, and the selection of the final configuration was based on the best trade-off between stable convergence and validation performance. An effective batch size of 32 samples and AdamW optimizer were used in the adopted configuration. The selected learning rate for full fine-tuning was $1\times10^{-5}$ for three epochs, whereas under LoRA, $1\times10^{-4}$ was used for four epochs. A cosine learning-rate schedule \cite{loshchilov2017sgdr} with a minimum value of 10\% of the initial rate was employed.

All SFT fine-tuning runs were executed in the following environment: OS \texttt{Linux-6.12.63-x86\_64-with-glibc2.35}, 24 CPU cores, 1 Nvidia H200 GPU, and 256\,GB of RAM.

\subsection{Direct Preference Optimization (DPO)}

Direct Preference Optimization (DPO) is a preference-alignment method that reparameterizes the RLHF problem to directly optimize a policy $\pi_{\theta}$ from preference pairs $\{(x, y_{w}, y_{\ell})\}$ without training a separate reward model \cite{rafailov2023dpo}. The DPO loss used in this research follows the original formulation:

\[
\mathcal{L}_{\mathrm{DPO}}(\pi_{\theta}; \pi_{\mathrm{ref}}) \;=\; -\mathbb{E}_{(x,y_w,y_\ell)\sim\mathcal{D}}\left[\log \sigma\!\left(\beta\left(\log\frac{\pi_{\theta}(y_w\mid x)}{\pi_{\mathrm{ref}}(y_w\mid x)} - \log\frac{\pi_{\theta}(y_\ell\mid x)}{\pi_{\mathrm{ref}}(y_\ell\mid x)}\right)\right)\right],
\]

where \(\sigma\) is the sigmoid function, \(\pi_{\text{ref}}\) is a fixed reference policy, and \(\beta\) is a scaling factor. The loss maximizes the log-probability margin between preferred output \(y_w\) and rejected output \(y_\ell\) relative to the reference policy, increasing the relative probability of desired outputs and decreasing that of undesired ones \cite{bradley1952rank}. In addition to its theoretical connection to the classic RLHF objective, DPO is computationally simple and empirically stable.

Although OCR is a sufficiently objective task that preference-based alignment methods have not, to the best of the authors' knowledge, been previously explored for it, the main motivation for using DPO in this research is the reduction in textual degeneration. SFT-tuned models can still exhibit degenerate behaviors that are not consistently removed by likelihood training alone. Here, DPO is used as a targeted refinement stage that explicitly pushes the policy away from those failure modes while preserving core transcription fidelity.

First, a pool consisting of the SFT evaluation set plus a random sample corresponding to 50\% of the SFT training instances, totaling \textbf{23\,726} samples, was defined for the construction of the preference dataset. Each sample was used for inference with four models previously selected for strong performance in the SFT stage. In addition to those four generations, the reference transcription was retained, resulting in five candidate responses per document.

All generated outputs were then evaluated by two large LLM judges, namely, Qwen3-VL-235B-A22B-Instruct \cite{qwen3_vl_235b_a22b_instruct} and gemma-3-27b-it \cite{gemma3_27b_it}, across the following four criteria: completeness, accuracy, formatting, and adherence to the expected JSON. Both judges assigned an integer score between 0 and 1000 for each criterion and provided a justification. Regarding calibration, all annotated inferences went through a 100-line warmup (initial replicas that were later discarded). A subsequent human qualitative analysis of the scores and the justifications indicated Qwen3-VL-235B-A22B-Instruct produced more faithful and consistent evaluations; therefore, only its annotations were retained for a subsequent pair-selection process. Details of the criteria and prompts are provided in Appendix \ref{apx:criterios-avaliacao}.

Preference pairs were constructed through exhaustive comparisons of the five responses within each document. All pairwise combinations were generated for each instance, yielding \(\binom{5}{2} = 10\) comparisons per document and 237\,260 candidate pairs across the corpus. The scores and the justifications provided by the selected judge were then used to guide pair selection.

A multi-stage filtering policy that combines principles from Selective-DPO \cite{gao2025principled} with analyses of DPO optimization dynamics \cite{xu2025beyond}, were applied to maximize instructional signal and avoid optimization conflicts due to problematic reward margins. Crucially, outputs displaying textual degeneration were deliberately preserved as \textit{rejected} examples whenever possible, providing a clean and consistent negative signal for the targeted behavior. Scores and justifications were further subjected to statistical analyses for the removal of outliers and biases, followed by a human review on random samples towards verification of coherence and calibration. Sensitivity analyses were also conducted over the adopted thresholds for assessment of robustness. After application of those selection rules and quality controls, the final preference dataset comprised 49\,170 pairs (approximately 21\% of the candidate pairs).

Inference generation and annotation were performed on an infrastructure with 8 Nvidia H200 GPUs and the final DPO training was executed on a cluster running \texttt{Linux-6.12.63-x86\_64-with-glibc2.35}, with 96 CPU cores, 4 Nvidia H200 GPUs, and 1024\,GB of RAM.

An effective batch size of 32, truncated inputs above 8\,192 tokens, and AdamW optimizer were used across all training runs and learning rates were explored in the $1\times10^{-5}$--$1\times10^{-6}$ range and trained for 3 to 5 epochs. The final configuration adopted was the one with best trade-off between loss stability and benchmark performance. The learning-rate schedule used cosine decay with a floor at 10\% of the initial rate and evaluations and checkpoints were executed periodically in each epoch.

\subsection{Quantization}

After training, the models with best benchmark evaluation metrics were quantized towards assessment of the way those metrics change and impact on computational performance and model cost. Model-quantization methods are numerical compression techniques for neural networks that reduce the precision of weights and/or activations to lower memory requirements and inference time while preserving model quality as much as possible \cite{wu2020integer_quantization, wei2024quantization_review}.

Activation-aware Weight Quantization (AWQ), a widely adopted state-of-the-art approach for LLM/SLM quantization that improves the quality--efficiency trade-off by accounting for activation outliers during weight quantization was used in this research \cite{lin2023awq}. The same corpus adopted for SFT training was used for the calibration of AWQ, but with a smaller data sample. The entire process was executed on the same hardware environment of the training, described in Subsection~\ref{subsec:SFT}.

Not all model parameters were quantized. The network output layer and the submodules in the model's visual branch (i.e., visual encoder and its associated submodules) were kept in full precision throughout the process. This choice is justified, since the output layer performs the final projection from hidden states to vocabulary logits and is particularly sensitive to perturbations introduced by quantization, a practice also adopted in reference implementations \cite{llmcompressor2024}. Similarly, the visual components of VLMs tend to require higher numerical precision for preserving stable multimodal representations, for the literature reports relevant sensitivity differences between visual tokens and linguistic tokens under quantization \cite{li2025mbq}.

AWQ-based quantization was evaluated under three precision configurations, chosen to span different quality--efficiency trade-offs, namely, AWQ FP8 (8-bit floating-point weights and activations), AWQ W8A8 (8-bit integer weights and activations), and AWQ W4A16 (4-bit integer weights with 16-bit activations). The activation formats were selected towards balancing kernel availability and numerical stability. FP8 provides a widely supported low-precision floating-point path for high throughput, W8A8 serves as a strong INT8 baseline, and W4A16 preserves higher-precision activations to reduce accuracy degradation when weights are aggressively quantized to 4 bits, since activation quantization is often more sensitive to outliers and dynamic-range effects \cite{lin2023awq, wei2024quantization_review}. The benchmark results obtained under those different configurations are analyzed in Appendix~\ref{apx:quant-benchmark-ocr}.

\section{Model evaluation methodology}
\label{sec:metodologia-avaliacao}

Large-scale OCR benchmarks such as \textit{OCRBench} \cite{liu2023ocrbench} and \textit{olmOCR-Bench} \cite{allenai2025olmocrbench} are not designed to reflect the constraints of the domain covered by this research, i.e., Brazilian Portuguese document OCR. In preliminary analyses, performance on these general-purpose benchmarks did not reliably measure performance in contexts where language-specific orthography, domain vocabulary, and document formatting can shift the error profile and amplify text degeneration. Those limitations motivated the proposal of a dedicated benchmark that captures the challenges of that domain and supports more actionable model comparisons.

\textit{DharmaOCR-Benchmark}, a dedicated benchmark that enables comparison of proprietary models, traditional OCR systems, and open-source models (specialized or general-purpose) in the proposed domain was, therefore, built. This section describes the methodology for evaluation of the models tested towards comparisons of different text-extraction approaches regarding transcription quality, unit inference cost, and thet degeneration rate. 

\subsection{DharmaOCR-Benchmark}

\textit{DharmaOCR-Benchmark} was designed to enable model comparisons in a specific domain, namely, text extraction in Brazilian Portuguese. The benchmark dataset is composed mainly of instances drawn from studies and from collected public datasets. The benchmark data were not used for training the evaluated models.

Of the 496 benchmark instances, 413 were sourced from previously established datasets from the literature and the remaining 83 were curated from publicly available documents, predominantly from the legal domain, towards increasing domain coverage. The sample breakdown is:

\begin{itemize}
  \item \textbf{ESTER-Pt (363 samples):} A public benchmark focused on Text Recognition in Brazilian Portuguese \cite{ester_pt}.
  \item \textbf{Legal (83 samples):} An internal set consisting of legal and administrative documents.
  \item \textbf{BRESSAY (50 samples):} A benchmark dedicated to handwritten text recognition \cite{neto2024bressay}.
\end{itemize}

A two-stage labeling pipeline inspired in olmOCR-Bench methodology \cite{poznanski2025olmocr} was implemented for the Legal subset, which, unlike the others, did not come from previously estabilished benchmarks. Labels were produced by multiple LLMs, followed by a complete manual audit of all instances.

A metric called \textbf{DharmaOCR-Benchmark Score}, was used for quantitative evaluation and defined as:

\begin{equation}
\label{eq:dharma-score}
score = \frac{Levenshtein_{Ratio} + BLEU}{2}
\end{equation}

where $Levenshtein_{Ratio}$ is a normalized edit-distance score derived from the Levenshtein distance \cite{levenshtein1966binary} computed between predicted transcription and ground-truth text and $BLEU$ (Bilingual Evaluation Understudy) is an n-gram overlap metric originally introduced for machine translation evaluation \cite{papineni2002bleu}.

In practice, $Levenshtein_{Ratio}$ captures fine-grained character-level fidelity (e.g., misspellings, missing accents, and punctuation errors), whereas $BLEU$ is more sensitive to preservation of local token sequences and penalizes severe re-orderings or dropped spans. Overall, those components are widely used to quantify transcription quality from complementary perspectives in document understanding and parsing settings (e.g., Levenshtein-style similarity \cite{kim2022donut,ouyang2024omnidocbench} and BLEU-style sequence fidelity \cite{hwang2023disgo}), and their aggregation provides a single scalar that reflects both aspects for model comparison.

Besides transcription quality, operational-viability metrics, namely, average text degeneration rate and unit cost per request were evaluated. The degeneration rate was computed by flagging requests that (i) hit the configured output-token limit and (ii) exhibit repeated text spans, detected with the use of an n-gram repetition criterion.

\subsection{Inference infrastructure}

The inference stack was standardized in an AWS environment so as to ensure reproducibility and economic feasibility. The benchmark was run for all models on g6e.2xlarge instances (NVIDIA L40S GPU with 48GB GDDR6, 8 AMD EPYC vCPUs, and 64 GiB of RAM) and execution orchestration was handled with vLLM, using a generated-context limit of 8,192 tokens for avoiding truncation on denser pages and complex tables.

\subsection{Cost calculation methodology}

Average cost per million pages metric was used for enabling a fair comparison between local models and commercial APIs. Cost for models run on local infrastructure, namely, fine-tuned models and open-source baselines, was estimated from the average processing time per page and the hourly cost of the infrastructure required to sustain execution.

Since the providers’ internal model cost cannot be computed for commercial models accessed via API, service prices were used for comparison. The computation of costs was based on public pricing tables valid in March 2026, by converting the provider billing unit to a million-page scale. For token-priced services (e.g., GPT-4o), the estimate is based on the average input and output tokens per page and the per-token prices, and for services with a fixed per-page fee (or per-extraction type), values are scaled proportionally. The formulas and the numeric breakdown (including the final values adopted for each API) are provided in Appendix~\ref{apx:calculo-custos}.

\section{Results}
\label{sec:results}

\subsection{Quantitative results}
\label{subsec:result_quant}

Table~\ref{tab:benchmark-ocr} summarizes the evaluation results for the models explored, including proprietary models, open-source baselines models, and variants trained. Since the cost of non-commercial models depends on the execution infrastructure, a relative cost is expressed as a percentage with respect to a reference model. \texttt{OlmOCR 2-7B}, which achieved the most consistent benchmark outcome (high score and low text degeneration rate), was chosen as the reference model.

\begin{table}[htbp]
\centering
\caption{Results of the models evaluated on DharmaOCR-Benchmark. Parentheses in the first column indicate the specialization techniques used. When a model is not indicated as LoRA, it means that full fine-tuning has been performed. Entries marked with ``Quant'' indicate AWQ-quantized variant with best performance among the quantized configurations, as detailed in Appendix~\ref{apx:quant-benchmark-ocr}.}
\begin{footnotesize}
{\setlength{\tabcolsep}{4pt}
\renewcommand{\arraystretch}{1.15}
\begin{tabularx}{\textwidth}{>{\raggedright\arraybackslash}X >{\centering\arraybackslash}m{1.4cm} >{\centering\arraybackslash}m{2.0cm} >{\centering\arraybackslash}m{2.0cm} >{\centering\arraybackslash}m{2.0cm}}
\toprule
\shortstack{\textbf{Model / Technique}} & \textbf{Score} & \shortstack{\textbf{Degeneration} \\ \textbf{rate (\%)}} & \shortstack{\textbf{Time per} \\ \textbf{page (s)}} & \shortstack{\textbf{Relative cost} \\ \textbf{(\%)}} \\
\midrule

\multicolumn{5}{l}{\textbf{Fine-tuned models}} \\

olmOCR-2-7B (SFT) & 0.928 & 1.01 & 2.717 & 63.09\% \\
olmOCR-2-7B (SFT + DPO) & 0.927 & 0.40 & 2.730 & 63.40\% \\
olmOCR-2-7B (LoRA SFT) & 0.926 & 0.60 & 2.997 & 69.60\% \\
olmOCR-2-7B (SFT + DPO + Quant) & 0.925 & 0.40 & 2.132 & 49.50\% \\
olmOCR-2-7B (Vanilla) & 0.823 & 1.41 & 4.306 & 100.00\% \\

\midrule

Nanonets-OCR2-3B (SFT + DPO) & 0.921 & 0.20 & 1.668 & 38.73\% \\
Nanonets-OCR2-3B (SFT) & 0.917 & 1.61 & 1.756 & 40.78\% \\
Nanonets-OCR2-3B (SFT + DPO + Quant) & 0.911 & 0.20 & 1.464 & 33.99\% \\
Nanonets-OCR2-3B (LoRA SFT) & 0.810 & 1.01 & 2.638 & 61.25\% \\
Nanonets-OCR2-3B (Vanilla) & 0.791 & 2.62 & 1.911 & 44.37\% \\

\midrule

Qwen2.5-VL-7B-Instruct (SFT + DPO) & 0.906 & 1.01 & 3.064 & 71.15\% \\
Qwen2.5-VL-7B-Instruct (SFT) & 0.904 & 1.61 & 2.861 & 66.44\% \\
Qwen2.5-VL-7B-Instruct (LoRA SFT) & 0.845 & 2.02 & 3.205 & 74.43\% \\
Qwen2.5-VL-7B-Instruct (Vanilla) & 0.839 & 2.42 & 3.101 & 72.01\% \\

\midrule

Qwen2.5-VL-3B-Instruct (SFT) & 0.819 & 3.23 & 1.881 & 43.69\% \\
Qwen2.5-VL-3B-Instruct (SFT + DPO) & 0.793 & 1.41 & 1.893 & 43.95\% \\
Qwen2.5-VL-3B-Instruct (LoRA SFT) & 0.786 & 4.03 & 2.258 & 52.44\% \\
Qwen2.5-VL-3B-Instruct (Vanilla) & 0.549 & 0.60 & 1.500 & 34.84\% \\

\midrule

gemma-3-4b-it (SFT + DPO) & 0.654 & 4.03 & 1.205 & 27.98\% \\
gemma-3-4b-it (SFT) & 0.640 & 16.13 & 1.430 & 33.21\% \\
gemma-3-4b-it (LoRA SFT) & 0.354 & 52.02 & 3.336 & 77.47\% \\
gemma-3-4b-it (Vanilla) & 0.214 & 33.96 & 2.182 & 50.68\% \\

\midrule
\multicolumn{5}{l}{\textbf{Open-source vanilla models}} \\

Qwen3-VL-8B-Instruct & 0.829 & 5.65 & 7.250 & 168.36\% \\
Dots OCR & 0.738 & 6.85 & 2.526 & 58.66\% \\
GLM-OCR & 0.710 & 11.69 & 1.480 & 34.36\% \\
Qwen3-VL-2B-Instruct & 0.623 & 11.69 & 3.566 & 82.80\% \\
DeepSeek-OCR & 0.196 & 21.98 & 1.213 & 28.16\% \\

\midrule
\multicolumn{5}{l}{\textbf{Commercial models}} \\

Claude-Opus-4.6 & 0.833 & -- & -- & -- \\
Gemini-3.1-Pro & 0.820 & -- & -- & -- \\
GPT-5.4 & 0.750 & -- & -- & -- \\
Google Vision & 0.686 & -- & -- & -- \\
Google Document AI & 0.640 & -- & -- & -- \\
GPT-4o & 0.635 & -- & -- & -- \\
Amazon Textract & 0.618 & -- & -- & -- \\
Mistral OCR 3 & 0.574 & -- & -- & -- \\

\bottomrule
\label{tab:benchmark-ocr}
\end{tabularx}}
\end{footnotesize}
\end{table}

DPO-aligned, quantized olmOCR2-7B achieved the third-highest benchmark score, trailing the top model by 0.003 (0.3\% of the top score) among all evaluated models. It also showed the second-lowest text degeneration rate and the lowest cost among models of same size.

DPO-aligned, quantized Nanonets-OCR2-3B attained a high average score among smaller 3B-parameter models, i.e., only 0.010 below the top model (approximately 1\% of the best score), the lowest degeneration rate across all models, and the lowest cost among the 3B models.

According to the results, those two models were selected as the best (jointly optimizing performance and cost) within their respective parameter regimes and are referred to as \textbf{DharmaOCR Full} and \textbf{DharmaOCR Lite}. The choice was motivated by their favorable trade-off among extraction quality, text degeneration rate, and cost within their capacity classes.

Compared to commercial systems and all open-source models, both DharmaOCR variants achieved the highest benchmark scores and the lowest text degeneration rates. Although they are not the absolute lowest-cost options (e.g., DeepSeek- OCR is cheaper), they provide the best overall balance and superior extraction quality at a highly competitive cost.

\begin{figure}[htbp]
    \centering
    \begin{tikzpicture}
    \begin{axis}[
        ybar,
        bar width=15pt,
        enlarge x limits=0.10,
        width=0.95\textwidth,
        height=7cm,
        ylabel={Text degeneration rate (\%)},
        symbolic x coords={
        gemma-3-4B-it,
        Qwen2.5-VL-7B,
        Qwen2.5-VL-3B,
        olmOCR-2-7B,
        Nanonets-OCR2-3B},
        xtick=data,
        nodes near coords,
        nodes near coords align={vertical},
        every node near coord/.append style={font=\tiny},
        legend style={
            at={(0.98,0.98)},
            anchor=north east,
            draw=black,
            fill=white,
            fill opacity=1,
            text opacity=1,
            font=\small,
            cells={anchor=west},
            inner sep=2pt,
        },
        legend image code/.code={\draw[#1] (0cm,-0.1cm) rectangle (0.28cm,0.1cm);},
        title={\textbf{Impact of tuning techniques on text degeneration}},
        ymajorgrids,
        grid style={dashed, gray!30},
    ]

    \addplot[fill=gray!40] coordinates {
    (Qwen2.5-VL-7B, 2.42)
    (Qwen2.5-VL-3B, 0.60)
    (olmOCR-2-7B, 1.41)
    (Nanonets-OCR2-3B, 2.62)
    (gemma-3-4B-it, 33.96)
    };
    \addlegendentry{Vanilla}

    \addplot[fill=violet!60] coordinates {
    (Qwen2.5-VL-7B, 1.61)
    (Qwen2.5-VL-3B, 3.23)
    (olmOCR-2-7B, 1.01)
    (Nanonets-OCR2-3B, 1.61)
    (gemma-3-4B-it, 16.13)
    };
    \addlegendentry{SFT}

    \addplot[fill=pink!80] coordinates {
    (Qwen2.5-VL-7B, 1.01)
    (Qwen2.5-VL-3B, 1.41)
    (olmOCR-2-7B, 0.40)
    (Nanonets-OCR2-3B, 0.20)
    (gemma-3-4B-it, 4.03)
    };
    \addlegendentry{SFT + DPO}

    \end{axis}
    \end{tikzpicture}

    \caption{Text degeneration rate (\%) across alignment stages. SFT reduces degeneration relative to \textit{vanilla} models in most cases, whereas DPO further reduces it, even compared to the SFT-tuned model.}
    \label{fig:degeneration_reduction}
\end{figure}
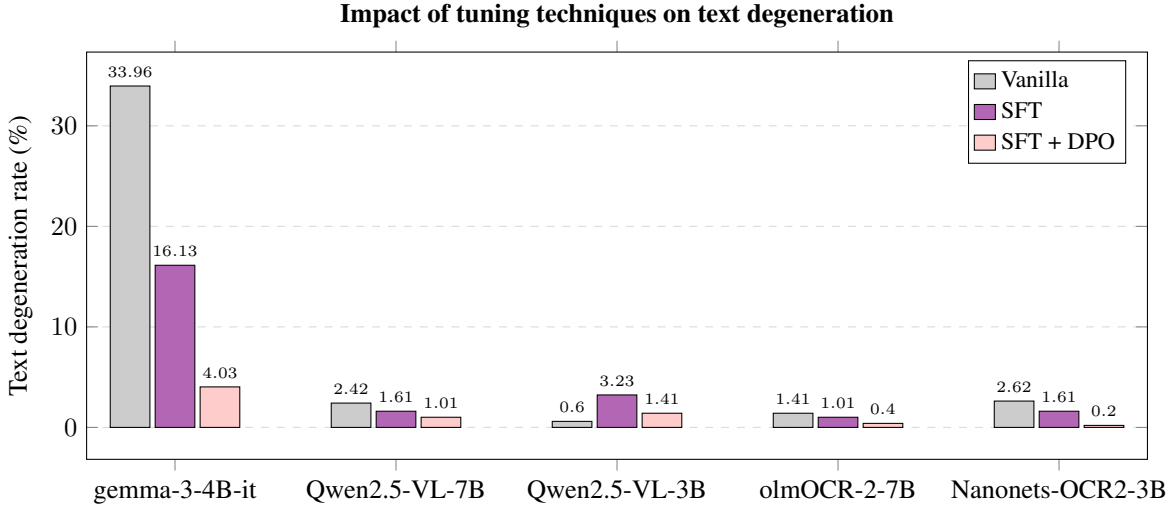

Another relevant aspect in Table~\ref{tab:benchmark-ocr} is the effect of DPO alignment. The application of DPO across models substantially reduced the text degeneration rate, as shown in Figure~\ref{fig:degeneration_reduction}. 

Quantization also has a relevant impact. The results indicate it reduces average inference time, hence, unit cost per request with no significant degradation in extraction quality. For instance, for quantized DharmaOCR Full model, unit cost was reduced by nearly 22\%, whereas the benchmark score decreased only slightly (by 0.002, approximately 0.2\% of the non-quantized model's score).

Overall, the results show the combination of the proposed specialization techniques substantially improves model quality, outperforming all open-source and commercial tested models. The resulting models achieved the lowest text degeneration rates among all evaluated systems and the unit inference cost of the best developed models was considerably reduced when compared to their vanilla counterparts.

\subsection{Performance--cost efficiency}

\begin{figure}[htbp]
    \centering
    \begin{tikzpicture}
    \begin{axis}[
        width=0.95\textwidth,
        height=8cm,
        xmode=log,
        log ticks with fixed point,
        xlabel={Cost per million pages (USD)},
        ylabel={DharmaOCR-Benchmark Score},
        xmin=500, xmax=65000,
        ymin=0.1, ymax=1.0,
        grid=both,
        grid style={line width=.1pt, draw=gray!10},
        major grid style={line width=.2pt, draw=gray!20},
        legend pos=south east,
        legend cell align={left},
        title={\textbf{Performance-to-Cost: DharmaOCR-Benchmark}},
        tick label style={font=\tiny},
        label style={font=\footnotesize},
        nodes near coords,
        every node near coord/.append style={
            font=\tiny\mdseries,
            anchor=south,
            yshift=2pt,
            /pgf/number format/fixed,
            check for zerfo/.code={
                \pgfkeys{/pgf/fpu}
                \pgfmathparse{\pgfplotspointmeta < 0.6}
                \ifnum\pgfmathresult=1\relax\pgfkeys{/pgf/node near coords/.style={}}\fi
                \pgfkeys{/pgf/fpu=false}
            }
        },
        point meta=explicit symbolic
    ]

    \addplot[
        only marks,
        mark=asterisk,
        mark size=4pt,
        green!50!black,
        mark options={line width=1pt},
        nodes near coords style={font=\scriptsize\bfseries, text=green!50!black},
    ] coordinates {
        (1327.76, 0.925) []
        (911.75, 0.911)  [DharmaOCR Lite]
    };
    \node[font=\scriptsize\bfseries, anchor=south, yshift=6pt, text=green!50!black]
        at (axis cs:1327.76,0.925) {DharmaOCR Full};
    \addlegendentry{DharmaOCR (Ours)}

    \addplot[
        only marks,
        mark=triangle*,
        mark size=2.5pt,
        violet,
    ] coordinates {
        (2681.68, 0.823) [olmOCR-2-7B]
        (1190.13, 0.791) [Nanonets-OCR2-3B]
        (1573.14, 0.738) [Dots OCR]
        (921.71,  0.710) [GLM-OCR]
        (755.43,  0.196) [DeepSeek-OCR]
    };
    \addlegendentry{Open Source Baselines Models}

    \addplot[
        only marks,
        mark=*,
        mark size=2.5pt,
        black,
    ] coordinates {
        (11475, 0.665) [Google Document AI]
        (9080, 0.635)  [GPT-4o]
        (20497.23, 0.750) [GPT-5.4]
        (47456.66, 0.833) [Claude-Opus-4.6]
        (14375.46, 0.820) [Gemini-3.1-Pro]
        (1500, 0.618)  [AWS Textract]
        (2000, 0.574)  [Mistral OCR 3]
        (1500, 0.686)  []
    };
    \node[font=\tiny\mdseries, anchor=south, yshift=0.5pt]
        at (axis cs:1500,0.680) {Google Vision};

    \addlegendentry{Commercial APIs}

    \end{axis}
    \end{tikzpicture}
    \caption{Quality–cost comparison among DharmaOCR models developed in this research, other open-source OCR models, and representative commercial APIs. Since the underlying model costs for commercial APIs are not available, their public prices were used as the comparison baseline.}
    \label{fig:performance_cost_labeled}
\end{figure}
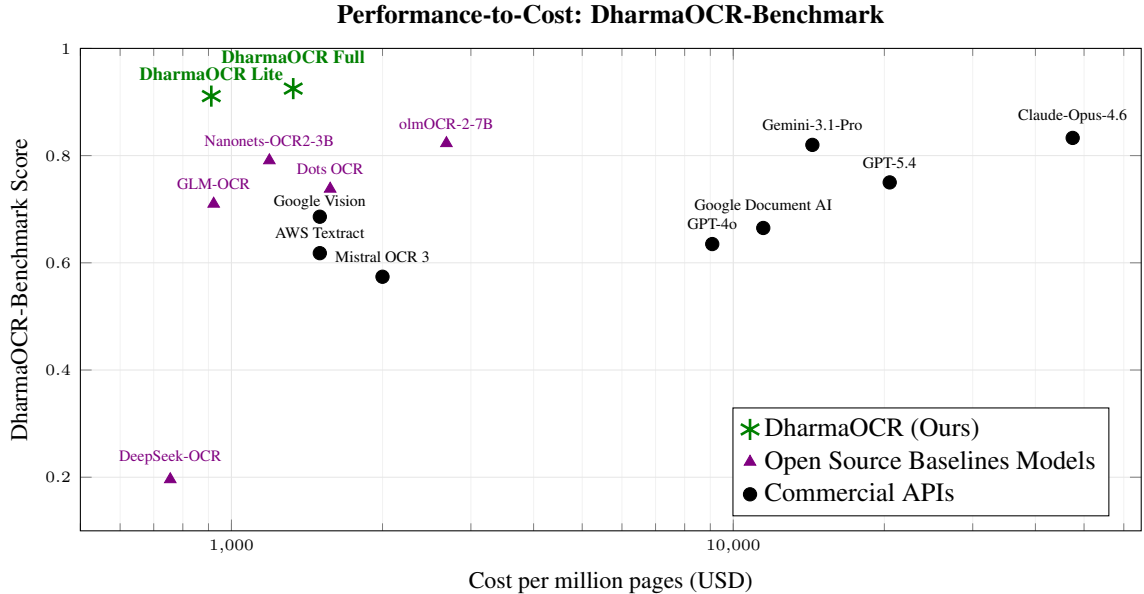

Model costs were computed (as detailed in Appendix~\ref{apx:calculo-custos}) using the hourly cost of an L40S GPU and the same vLLM inference configuration for all locally executed models for ensuring comparability. Figure~\ref{fig:performance_cost_labeled} displays the relationship between cost and quality for the evaluated models.

The models developed in this research dominated the cost–benefit Pareto frontier, outperforming commercial systems and other open-source OCR-specialized models. Notably, they also surpassed general-purpose, state-of-the-art LLM baselines included in the proposed benchmark (e.g., Claude-Opus-4.6, Gemini-3.1-Pro, and GPT-5.4)at a much lower cost. Note the cost axis in Fig.~\ref{fig:performance_cost_labeled} is logarithmic – as an example, DharmaOCR Lite is approximately 52 times cheaper than Claude-Opus-4.6 and achieves higher benchmark quality.

A caveat is the estimated cost of running open-source models locally was compared against the public prices of commercial APIs, which can be disadvantageous to commercial providers. However, local execution costs are strongly influenced by GPU choice and vLLM configuration used for optimizing hardware utilization.

Accordingly, model-specific inference settings on newer GPUs can significantly reduce costs. Towards illustrating it, the benchmark was re-run on DharmaOCR Full by an H200 GPU with a carefully tuned inference configuration.

A single-H200 pod carved out of an AWS \texttt{p5e.48xlarge} instance (8\,$\times$\,H200) priced at USD 39.80 per hour. was the infrastructure used for the test. Under a Kubernetes setup, the full machine was split into eight identical pods, each with one H200, and the experiment was run on one of them. Therefore, the hourly infrastructure cost for the test was computed as the full-instance cost divided by 8, i.e., $\frac{39.8}{8} = 4.975$ USD/hour.

Under the same protocol adopted on the L40S machine, the test yielded a substantial reduction in both average time per page, which decreased from 2.1\,s to 0.57\,s (nearly 4$\times$ lower), and cost per million pages, wihch dropped from USD 1327.76 to USD 787.70 (nearly 2$\times$ lower).

Overall, the results indicate DharmaOCR outperformed all other solutions evaluated on that benchmark, achieving markedly higher quality at a competitive cost (with further room for cost reductions).

\subsection{LLM-as-a-judge qualitative evaluation}

In addition to using DharmaOCR-Benchmark for model comparison, a qualitative evaluation was conducted with an LLM-as-a-judge setup, with GPT-4o as the judge, for comparing the main models developed and validating the results reported in the previous sections.

Judge LLM selected the best extraction between two anonymized model outputs for each page, according to predefined criteria. The criteria and the evaluation procedure are described in Appendix~\ref{apx:llm-as-judge}.

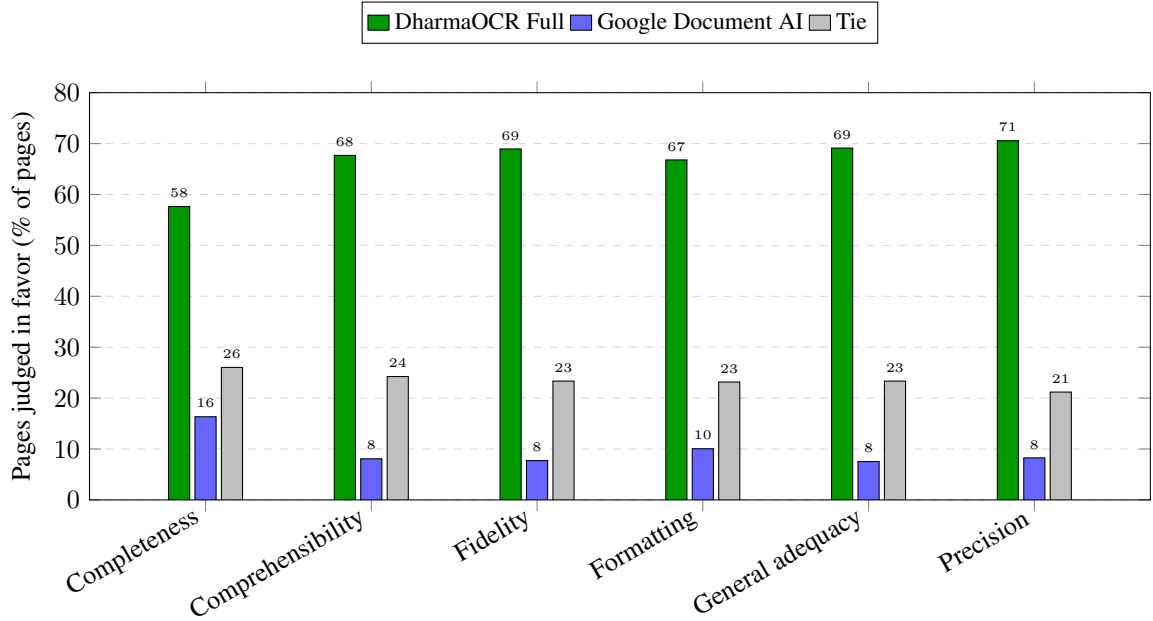
\begin{figure}[htbp]
    \centering
    \begin{tikzpicture}
    \begin{axis}[
        ybar,
        bar width=8pt,
        enlarge x limits=0.14,
        width=0.95\textwidth,
        height=7cm,
        ymin=0,
        ymax=80,
        ylabel={Pages judged in favor (\% of pages)},
        ytick={0,10,...,80},
        symbolic x coords={Completeness,Comprehensibility,Fidelity,Formatting,General\ adequacy,Precision},
        xtick=data,
        x tick label style={rotate=30, anchor=east},
        point meta=y,
        nodes near coords={\pgfmathprintnumber[fixed,precision=0]{\pgfplotspointmeta}},
        nodes near coords align={vertical},
        every node near coord/.append style={font=\fontsize{5}{6}\selectfont},
        legend style={
            at={(0.5,1.12)},
            anchor=south,
            legend columns=3,
            draw=black,
            fill=white,
            fill opacity=1,
            text opacity=1,
            font=\small,
            cells={anchor=west},
            inner sep=2pt,
        },
        legend image code/.code={\draw[#1] (0cm,-0.1cm) rectangle (0.28cm,0.1cm);},
        ymajorgrids,
        grid style={dashed, gray!30},
    ]

    \addplot[fill=green!60!black] coordinates {
        (Completeness,57.630162)
        (Comprehensibility,67.684022)
        (Fidelity,68.940754)
        (Formatting,66.786355)
        (General\ adequacy,69.120287)
        (Precision,70.556553)
    };
    \addlegendentry{DharmaOCR Full}

    \addplot[fill=blue!60] coordinates {
        (Completeness,16.337522)
        (Comprehensibility,8.078995)
        (Fidelity,7.719928)
        (Formatting,10.053860)
        (General\ adequacy,7.540395)
        (Precision,8.258528)
    };
    \addlegendentry{Google Document AI}

    \addplot[fill=gray!50] coordinates {
        (Completeness,26.032316)
        (Comprehensibility,24.236984)
        (Fidelity,23.339318)
        (Formatting,23.159785)
        (General\ adequacy,23.339318)
        (Precision,21.184919)
    };
    \addlegendentry{Tie}

    \end{axis}
    \end{tikzpicture}
    \caption{LLM-as-a-judge results from a comparison between DharmaOCR Full and Google Document AI. Bars show the percentage of pages in which the judge gave preference to each model or declared a tie for each evaluation criterion.}
    \label{fig:full_vs_documentai}
\end{figure}

The evaluation aimed not to perform exhaustive pairwise comparisons across all models, but to validate key patterns observed in the benchmark results. First, DharmaOCR’s advantage over commercial models was assessed by comparing DharmaOCR Full against DocumentAI, a widely used commercial system whose document understanding pipeline is also empowered by LLM-style generative models, making it a strong LLM-based commercial baseline. Figure~\ref{fig:full_vs_documentai} provides the comparison results.

\begin{figure}[htbp]
    \centering
    \begin{tikzpicture}
    \begin{axis}[
        ybar,
        bar width=8pt,
        enlarge x limits=0.14,
        width=0.95\textwidth,
        height=7cm,
        ymin=0,
        ymax=60,
        ylabel={Pages judged in favor (\% of pages)},
        ytick={0,10,...,60},
        symbolic x coords={Completeness,Comprehensibility,Fidelity,Formatting,General\ adequacy,Precision},
        xtick=data,
        x tick label style={rotate=30, anchor=east},
        point meta=y,
        nodes near coords={\pgfmathprintnumber[fixed,precision=0]{\pgfplotspointmeta}},
        nodes near coords align={vertical},
        every node near coord/.append style={font=\fontsize{5}{6}\selectfont},
        legend style={
            at={(0.5,1.12)},
            anchor=south,
            legend columns=3,
            draw=black,
            fill=white,
            fill opacity=1,
            text opacity=1,
            font=\small,
            cells={anchor=west},
            inner sep=2pt,
        },
        legend image code/.code={\draw[#1] (0cm,-0.1cm) rectangle (0.28cm,0.1cm);},
        ymajorgrids,
        grid style={dashed, gray!30},
    ]

    \addplot[fill=green!60!black] coordinates {
        (Completeness,46.774194)
        (Comprehensibility,54.838710)
        (Fidelity,51.612903)
        (Formatting,49.596774)
        (General\ adequacy,53.024194)
        (Precision,51.209677)
    };
    \addlegendentry{DharmaOCR Full}

    \addplot[fill=violet!45] coordinates {
        (Completeness,12.298387)
        (Comprehensibility,12.500000)
        (Fidelity,14.314516)
        (Formatting,9.072581)
        (General\ adequacy,13.306452)
        (Precision,14.516129)
    };
    \addlegendentry{olmOCR-2-7B}

    \addplot[fill=gray!50] coordinates {
        (Completeness,40.927419)
        (Comprehensibility,32.661290)
        (Fidelity,34.072581)
        (Formatting,41.330645)
        (General\ adequacy,33.669355)
        (Precision,34.274194)
    };
    \addlegendentry{Tie}

    \end{axis}
    \end{tikzpicture}
    \caption{LLM-as-a-judge results from a comparison between DharmaOCR Full and olmOCR-2-7B. Bars show the percentage of pages in which the judge gave preference to each model or declared a tie for each evaluation criterion.}
    \label{fig:full_vs_olm}
\end{figure}
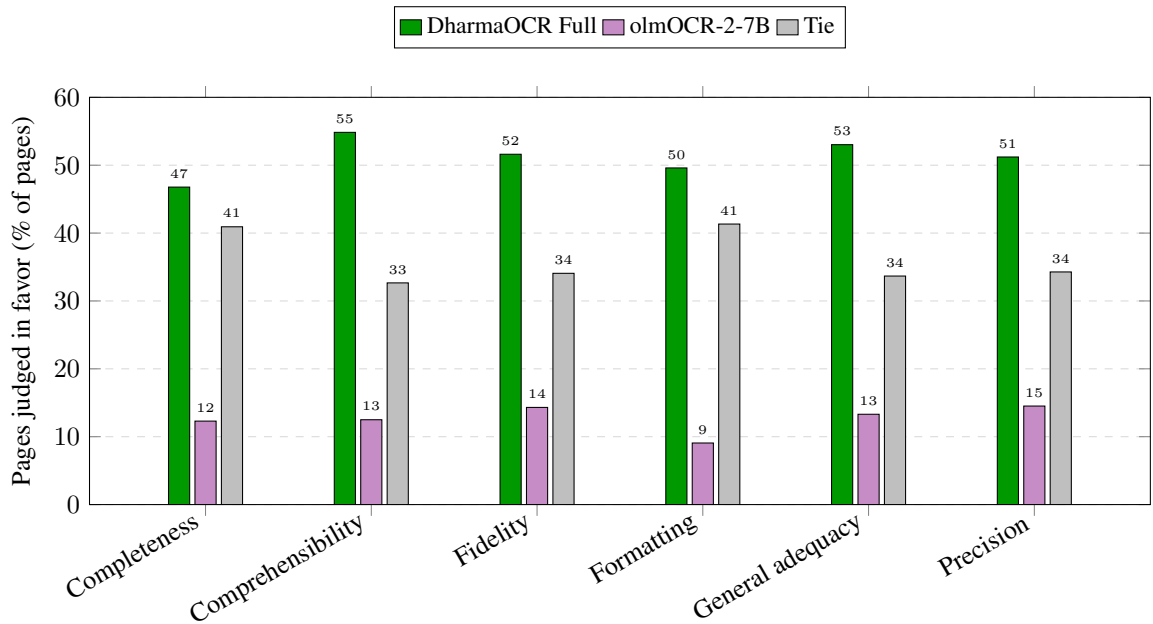

The plot shows LLM judge assigns more wins to DharmaOCR Full across all criteria, by a wide margin; notably, the tie rate is higher than DocumentAI's win rate. Furthermore, DharmaOCR Full was also compared to the benchmark's best-performing open-source model, OlmOCR-2-7B (Section~\ref{subsec:result_quant}). See Figure~\ref{fig:full_vs_olm}.

\begin{figure}[htbp]
    \centering
    \begin{tikzpicture}
    \begin{axis}[
        ybar,
        bar width=8pt,
        enlarge x limits=0.14,
        width=0.95\textwidth,
        height=7cm,
        ymin=0,
        ymax=70,
        ylabel={Pages judged in favor (\% of pages)},
        ytick={0,10,...,70},
        symbolic x coords={Completeness,Comprehensibility,Fidelity,Formatting,General\ adequacy,Precision},
        xtick=data,
        x tick label style={rotate=30, anchor=east},
        point meta=y,
        nodes near coords={\pgfmathprintnumber[fixed,precision=0]{\pgfplotspointmeta}},
        nodes near coords align={vertical},
        every node near coord/.append style={font=\fontsize{5}{6}\selectfont},
        legend style={
            at={(0.5,1.12)},
            anchor=south,
            legend columns=3,
            draw=black,
            fill=white,
            fill opacity=1,
            text opacity=1,
            font=\small,
            cells={anchor=west},
            inner sep=2pt,
        },
        legend image code/.code={\draw[#1] (0cm,-0.1cm) rectangle (0.28cm,0.1cm);},
        ymajorgrids,
        grid style={dashed, gray!30},
    ]

    \addplot[fill=orange!55] coordinates {
        (Completeness,18.850987)
        (Comprehensibility,20.107720)
        (Fidelity,20.825853)
        (Formatting,16.337522)
        (General\ adequacy,20.825853)
        (Precision,20.287253)
    };
    \addlegendentry{DharmaOCR Lite}

    \addplot[fill=green!60!black] coordinates {
        (Completeness,25.673250)
        (Comprehensibility,40.215440)
        (Fidelity,41.651706)
        (Formatting,19.569120)
        (General\ adequacy,42.010772)
        (Precision,41.831239)
    };
    \addlegendentry{DharmaOCR Full}

    \addplot[fill=gray!50] coordinates {
        (Completeness,55.475763)
        (Comprehensibility,39.676840)
        (Fidelity,37.522442)
        (Formatting,64.093357)
        (General\ adequacy,37.163375)
        (Precision,37.881508)
    };
    \addlegendentry{Tie}

    \end{axis}
    \end{tikzpicture}
    \caption{LLM-as-a-judge results from a comparison between DharmaOCR Full and DharmaOCR Lite. Bars show the percentage of pages in which the judge gave preference to each model or declared a tie for each evaluation criterion.}
    \label{fig:full_vs_lite}
\end{figure}
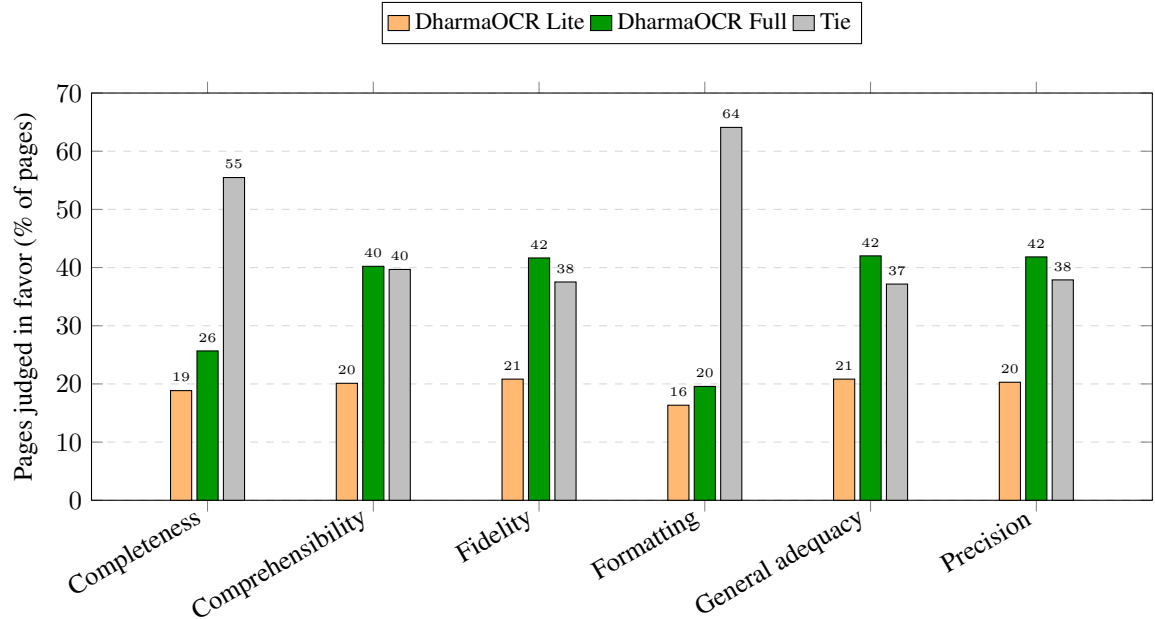

The plot further indicates a strong superiority of DharmaOCR Full over the best-performing open-source OCR model in the benchmark, outperforming olmOCR-2-7B across all evaluated criteria. Finally, it was also compared against DharmaOCR Lite (see Figure ~\ref{fig:full_vs_lite} for the results).

Consistently with the benchmark score, the plot shows DharmaOCR Full outperformed DharmaOCR Lite with a smaller margin than in comparisons with other models, but with an advantage in all criteria. Notably, the tie rate is higher, indicating a closer match between the models.

The results, obtained via a GPT-4o-assisted qualitative evaluation, help confirm the benchmark findings and support the validity of the adopted score, since all conclusions drawn from the LLM-as-a-judge comparisons are consistent with the results in Section~\ref{subsec:result_quant}:

\begin{itemize}
    \item DharmaOCR Full's extraction quality is substantially higher than that of leading commercial and open-source OCR models;
    \item DharmaOCR Full also outperforms olmOCR-2-7B, but the quality gap in favor of DharmaOCR Full is larger against DocumentAI than against olmOCR-2-7B;
    \item DharmaOCR Full outperforms DharmaOCR Lite, but by a smaller margin in comparison to the other models, suggesting DharmaOCR Lite should also outperform the two external baselines.
\end{itemize}

\section{Discussion}

\subsection{Specialization for efficiency optimization}

The results in Section~\ref{sec:metodologia-avaliacao} show specializing SLMs for OCR in a specific domain yields substantial gains not only over vanilla multimodal models, but also over models previously specialized for general OCR and established commercial solutions, supporting the claim contextual specialization can be more decisive than number of model parameters alone. Models used in their vanilla form, or with only partial specialization, perform substantially worse than versions specialized to the target domain.

Concretely, specialization enables quantized DharmaOCR Full to reach a $0.925$ benchmark score (Table~\ref{tab:benchmark-ocr}), outperforming similarly sized vanilla multimodal models such as Qwen2.5-VL-7B-Instruct (0.839) and Qwen3-VL-8B-Instruct (0.829), and a general-OCR specialized model in its base form, olmOCR-2-7B vanilla (0.823), by roughly 10--12\% in relative score. Text degeneration rate shows an approximately 3 to 14 times reduction compared to those baselines (e.g., from 1.41\% to 0.40\% for olmOCR-2-7B and from 5.65\% to 0.40\% for Qwen3-VL-8B-Instruct).

The benchmark score for quantized DharmaOCR Lite reaches $0.911$, substantially surpassing smaller vanilla models such as Qwen3-VL-2B-Instruct (0.623) and Qwen2.5-VL-3B-Instruct (0.549), and Nanonets-OCR2-3B vanilla baseline (0.791), with roughly 46--66\% relative gains over vanilla models and around 15\% over Nanonets vanilla. Regarding stability, the degeneration rate drops from 2.62\% to 0.20\% relative to Nanonets vanilla (approximately 13 times reduction) and from 0.60\% to 0.20\% relative to Qwen2.5-VL-3B-Instruct vanilla (3 times reduction).

Smaller models(e.g., quantized DharmaOCR Lite (~3B parameters)) specialized for OCR in the target domain outperform larger models also specialized for OCR, but in a general domain (e.g., olmOCR-2-7B vanilla \cite{poznanski2025olmocr2unittest} (with more than twice as many parameters)), by roughly 10\% in benchmark score.

Smaller and quantized models with no quality loss (with quality gains) reduce unit cost per page while increasing throughput. As an example, quantized DharmaOCR Lite not only achieves higher quality than olmOCR-2-7B vanilla, but also reduces average time per page from 4.306\,s to 1.464\,s and relative cost by approximately 67\% (Table~\ref{tab:benchmark-ocr}). Moreover, DharmaOCR models show consistently lower text degeneration rates than non-specialized models or models specialized for a general domain.

When compared to widely used commercial models, specialization is even more pronounced. Proprietary solutions such as Amazon Textract (0.618), Google Document AI (0.640), and Google Vision (0.686) achieve benchmark scores below those of proposed models (0.925 for quantized DharmaOCR Full and 0.911 for quantized DharmaOCR Lite), corresponding to approximately 35\% relative gains over Google Vision and 50\% over Amazon Textract, and also surpassing proprietary LLMs used as OCR (e.g., Claude-Opus-4.6 (0.833), Gemini-3.1-Pro (0.820), GPT-5.4 (0.750), and GPT-4o (0.635)). Therefore, DharmaOCR not only outperforms commercial models in the evaluated domain, but also offers a more computationally and economically efficient solution for high-volume scenarios.

\subsection{Incremental specialization}

A comparison of specializations built on top of vanilla models versus models previously specialized (Table~\ref{tab:benchmark-ocr}) (e.g., additional specialization on olmOCR-2-7B versus Qwen2.5-VL-7B), which share the same base architecture, shows a cumulative distribution-alignment effect. Specializing a model, already OCR-oriented tends to yield better final performance, suggesting starting closer to the target task distribution facilitates adaptation to the new domain. This is evident when tuned models based on olmOCR-2-7B versus Qwen2.5-VL-7B, and tuned models based on Nanonets-OCR2-3B versus Qwen2.5-VL-3B are compared, since the compared pairs have identical architectures.

In the first comparison, the best model derived from Qwen2.5-VL-7B-Instruct (SFT+DPO) reaches 0.906 with a 1.01\% text degeneration rate, whereas the best model based on olmOCR-2-7B (SFT+DPO) reaches 0.927 with 0.40\% degeneration, an approximately 2.3\% gain in quality score and a nearly 2.5 times reduction in degeneration rate.

In the second comparison, the gap is even larger, i.e., the best model derived from Qwen2.5-VL-3B-Instruct (SFT+DPO) reaches 0.793 with 1.41\% degeneration and the best model based on Nanonets-OCR2-3B (SFT+DPO) reaches 0.921 with 0.20\% degeneration, an approximately 16\% gain in quality score and nearly a 7 times reduction in degeneration rate.

The results indicate incremental specialization benefits from an OCR-aligned starting point, producing consistent gains in both quality and stability, thus suggesting multiple specialization levels may be defined for generative models in function of task specificity, i.e., general-domain vanilla open-source models (e.g., Qwen and Gemma families) as a first level, with subsequent levels emerging as deeper specialization is required for particular tasks and domains.

\begin{figure}
    \centering
    \includegraphics[width=\linewidth]{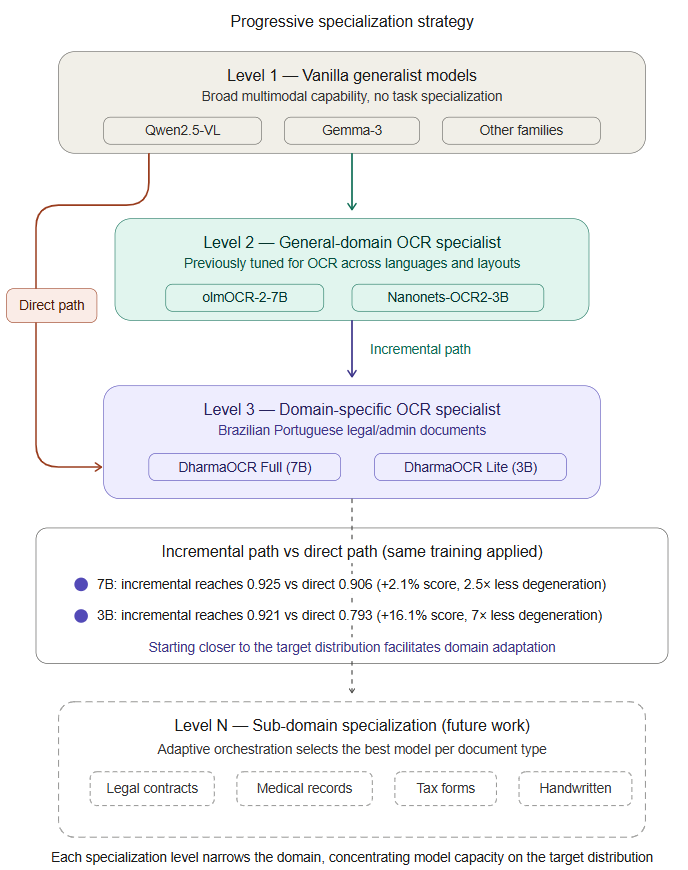}
    \caption{Progressive specialization strategy and comparison of two training paths.  Three specialization levels are shown---vanilla generalist (Level~1), general-domain OCR specialist (Level~2), and domain-specific OCR specialist (Level~3)---plus a projected Level~$N$ for future sub-domain specialization.}
    \label{fig:progressive-specialization}
\end{figure}

Figure~\ref{fig:progressive-specialization} formalizes that observation as a progressive specialization strategy. Rather than treating each fine-tuning experiment in isolation, the results indicate models absorb additional domain-specific alignment more effectively when they start closer to the target distribution. 

The diagram also contrasts two specialization paths that depart from the same Level~1 base architecture.  The direct path applies domain-specific SFT\,+\,DPO directly to the vanilla model (Level~1\,$\to$\,Level~3), whereas the incremental path first passes through a general-OCR stage before applying the same domain-specific training (Level~1\,$\to$\,Level~2\,$\to$\,Level~3). The incremental path consistently outperforms the direct one for both 7B and 3B architectures and the gains are obtained with the same domain-specific training procedure applied to both paths - the only difference is the starting point.

That pattern suggests a replicable specialization pipeline in which each level progressively narrows the domain and concentrates model capacity on the target distribution. Since each level builds on the distributional gains of the previous one, the strategy is inherently extensible, i.e., practitioners can develop new Level~$N$ specialists for other, more specific domains, such as medical records, tax forms, or legal contracts.

\subsection{DPO for reducing text degeneration}

One of the most relevant findings of this research is the text degeneration reduction throughout the specialization pipeline, particularly via DPO. As discussed in Section~\ref{sec:degeneracao-textual} and empirically demonstrated in Appendix~\ref{apx:impacto-degeneracao}, text degeneration affects not only output quality, but also inference-system efficiency, significantly increasing unit cost and average latency per request.

According to Table~\ref{tab:tempo_degeneracao} shows that, besides throughput reduction, replacing degenerate requests with average-duration requests might reduce total execution time by more than 42.47\% in some scenarios, indicating a direct computational-cost impact. Moreover, those estimates do not include costs of practical mitigation strategies such as real-time degeneration detection and per-occurrence retries. In light of those results, the fraction of requests with text degeneration is not only a key quality metric, but also a highly relevant operational metric.

A comparison of models trained only with SFT against those subsequently aligned with DPO shows the proposed use of DPO for reducing text degeneration is effective, since all models display a substantial reduction in degeneration rate. The magnitude of the reduction varies according to model, as shown in Table~\ref{tab:reducao-degeneracao-dpo}.

\begin{table}[ht]
\centering
{\setlength{\tabcolsep}{10pt}
\renewcommand{\arraystretch}{1.35}
\begin{tabular}{>{\raggedright\arraybackslash}m{3.2cm} >{\centering\arraybackslash}m{3.2cm} >{\centering\arraybackslash}m{3.6cm} >{\centering\arraybackslash}m{3.0cm}}
\hline
\rule{0pt}{6.0ex} \shortstack{\textbf{Model} \\\vphantom{rate (\%)}} & \shortstack{\textbf{SFT degeneration}\\\textbf{rate (\%)}} & \shortstack{\textbf{SFT+DPO degeneration}\\\textbf{rate (\%)}} & \shortstack{\textbf{Relative}\\\textbf{reduction}} \\
\hline
olmOCR-2-7B & 1.01 & 0.60 & 40.6\% \\
Nanonets-OCR2-3B & 1.61 & 0.20 & 87.6\% \\
Qwen2.5-VL-7B & 1.61 & 1.01 & 37.3\% \\
Qwen2.5-VL-3B & 3.23 & 1.41 & 56.4\% \\
gemma-3-4b-it & 16.13 & 4.03 & 75.0\% \\
\hline
\end{tabular}
}
\caption{Comparison of text degeneration rate between models tuned only with SFT and those tuned with SFT followed by DPO.}
\label{tab:reducao-degeneracao-dpo}
\end{table}

The reduction in text degeneration rate after application of DPO is substantiall, i.e., on average, a 59.4\% relative reduction, reaching 87.6\% for Nanonets-OCR2-3B (from 1.61\% to 0.20\%), a roughly 8 times decrease in occurrences compared to SFT-only tuning.

The results are consistent with the discussion in Section~\ref{sec:degeneracao-textual}. As shown in \cite{holtzman2020curiouscaseneuraltext,Yao_2025}, degeneration can occur when a model falls into a self-reinforcing probabilistic attractor in a high-likelihood region. Therefore, SFT, which maximizes conditional likelihood autoregressively, may still induce such attractors even after specialization.

In contrast, DPO penalizes complete rejected responses and rewards complete accepted ones. When preference pairs include degenerate examples labeled as rejected, DPO imposes an explicit negative probability shift upon those regions of the distribution, while remaining close to the SFT reference policy so as not to lose what was learned in the previous stage \cite{rafailov2023dpo}.

That interpretation is also consistent with evidence that pure maximum-likelihood objectives tend to over-assign probability to repetitive patterns and that adding training terms that explicitly penalize undesired continuations reduces repetition and looping while preserving overall quality \cite{welleck2019neural}. Therefore, DPO can be seen as a preference-guided form of ``implicit unlikelihood,'' allowing the model to learn semantic and behavioral negatives, such as text degeneration, from rejected examples.

\section{Conclusions and Future Work}

This manuscript introduced DharmaOCR Full and Lite, two Specialized Small Language Models (SSLMs) for structured OCR, with a simultaneous focus on transcription quality, generation stability, and computational efficiency. Unlike traditional OCR approaches or large-scale multimodal models used in a generic manner, it showed targeted specialization of SLMs for a specific domain yields substantial gains in both qualitative performance and operational metrics.

The proposed pipeline combines multiple training techniques. First, Supervised Fine-Tuning (SFT) is applied for adapting to the domain and improving the structural adherence of JSON-formatted outputs (with header, text, footer, and margin fields). An additional alignment stage is then applied to OCR models for the first time, to the best of the authors’ knowledge, via Direct Preference Optimization (DPO) for reducing the text degeneration rate. After training, AWQ quantization is applied at different precision levels towards increasing throughput and reducing unit cost per request, with minimal quality loss.

As empirically demonstrated, degeneration is not only an output-quality problem, but also a critical driver of cost and throughput in real inference systems and should therefore be tracked in benchmarks for generative models. The use of DPO as an explicit mechanism to penalize degenerative behavior proved effective, significantly reducing degeneration and improving operational reliability.

Additionally, the manuscript presented DharmaOCR-Benchmark, a domain-specific benchmark (Brazilian Portuguese) that simultaneously evaluates textual fidelity, structural preservation, and generation stability. According to the results, DharmaOCR models outperformed vanilla open-source models and other OCR-tuned models and also surpassed all evaluated commercial solutions in both extraction quality and operational metrics, with lower text degeneration rates and substantially reduced cost per million processed pages.

Overall, the results support the central hypothesis that contextual specialization of smaller models can be more decisive than simply increasing the number of model parameters. When properly aligned to the target domain and expected output format, smaller models can surpass larger general-purpose ones in quality while substantially reducing cost and increasing throughput.

\subsection{Limitations and future work}

Despite strong results, several research directions remain open for further improvements to the models.

First, the model shows a tendency to repeat parts of the header, footer, or margins within JSON text field. Towards mitigating it in production, a post-processing step that checks exact matches between those fields and the beginning or end of the text was added. It solved more than 80\% of occurrences, but left room for improvements.

Therefore, the adoption of Reinforcement Learning with Verifiable Rewards (RLVR) is planned for structural validation of JSON outputs. The idea is to use verifiable rewards based on the checking of whether a repeated text appears across the expected schema fields, encouraging structurally separable outputs.

Second, in a broader research direction, a study of text degeneration in SLMs is intended towards systematically understanding the mechanisms that produce such behaviors and exploring additional mitigation strategies besides those adopted in this study.

Finally, specialization strategies will be investigated for different document types (e.g., legal, administrative, and handwritten ones, as well as forms and invoices) combined with automatic OCR orchestration by document type, which might lead to an adaptive system that dynamically selects the most appropriate model for each document category, maximizing quality and efficiency.

Such directions are expected to contribute to not only the evolution of DharmaOCR models, but also the broader progress in both understanding efficient specialization of small language models in real-world applications and reducing text degeneration, a general issue in autoregressive generative models.

\bibliographystyle{unsrt}
\bibliography{main}

\appendix
\clearpage
\section{Appendix}

\subsection{Impact of text degeneration on system performance}
\label{apx:impacto-degeneracao}

Qwen2.5-VL-7B-Instruct \cite{qwen2p5_vl_7b} was served with the use of vLLM for evaluation of the impact of text degeneration on system performance. Three OCR datasets, namely, DharmaOCR-Benchmark and two proprietary datasets with varied structures were used.

Requests from each dataset were issued by a semaphore of size 30, allowing at most 30 concurrent requests to run in parallel towards balancing latency and throughput. Table~\ref{tab:config-vllm-degeneracao} provides the vLLM settings - unspecified parameters follow vLLM v0.15.0 defaults \cite{vllm2026}. The machine used for inference was the same that of the model-evaluation setup described in Section~\ref{sec:metodologia-avaliacao}.

\begin{table}[htbp]
\centering
\caption{vLLM configuration used for the inference runs in the text degeneration analysis.}
\label{tab:config-vllm-degeneracao}
\begin{tabular}{ll}
\toprule
\textbf{Parameter} & \textbf{Value} \\
\midrule
\texttt{gpu\_memory\_utilization} & 0.90 \\
\texttt{tensor\_parallel\_size}   & 1 \\
\texttt{max\_num\_batched\_tokens} & 32000 \\
\texttt{max\_model\_len}          & 65536 \\
\texttt{temperature}              & 0 \\
\texttt{max\_tokens}              & 8192 \\
\bottomrule
\end{tabular}
\end{table}

During execution, both start and end time of each request were recorded. Degenerate requests were identified as those that hit the maximum generated-token limit and, upon human inspection, exhibited a word-repetition pattern at the end of the output. Requests without that pattern were considered healthy. The plots in Figures~\ref{fig:req_tempo_conjunto1}, \ref{fig:req_tempo_conjunto2}, and~\ref{fig:req_tempo_conjunto3}, which show when each request started and finished, highlighting degenerate cases in red, were produced with the use of those logs.

\afterpage{%
\begin{center}
  \includegraphics[height=0.92\textheight]{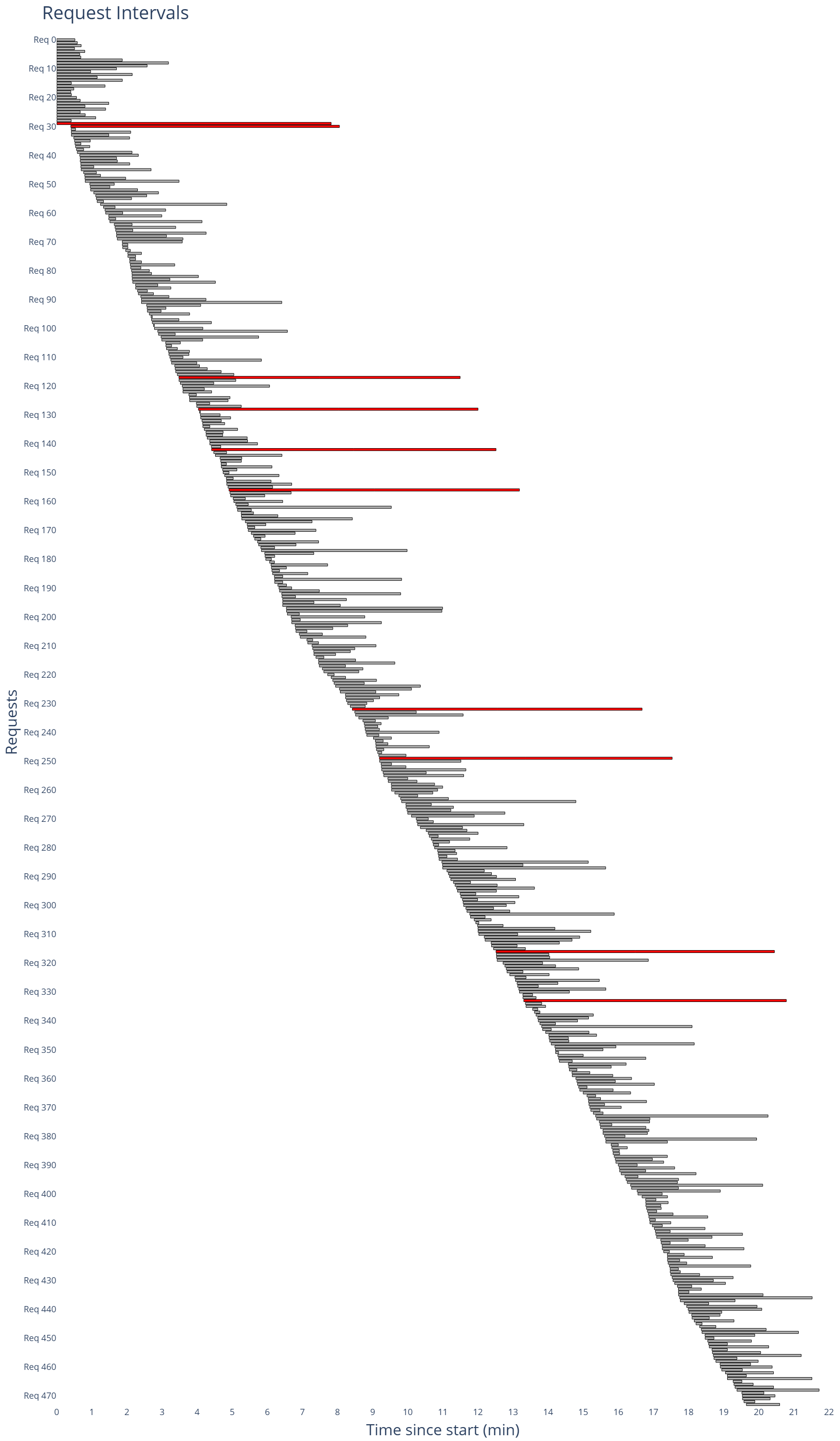}
  \captionof{figure}{Start and end time of each request (in submission order) for dataset 1. Each request is represented by a bar whose left edge marks start time and whose right edge marks end time. Degenerate requests are highlighted in red.}
  \label{fig:req_tempo_conjunto1}
\end{center}
\newpage}

\afterpage{%
\begin{center}
  \includegraphics[height=0.92\textheight]{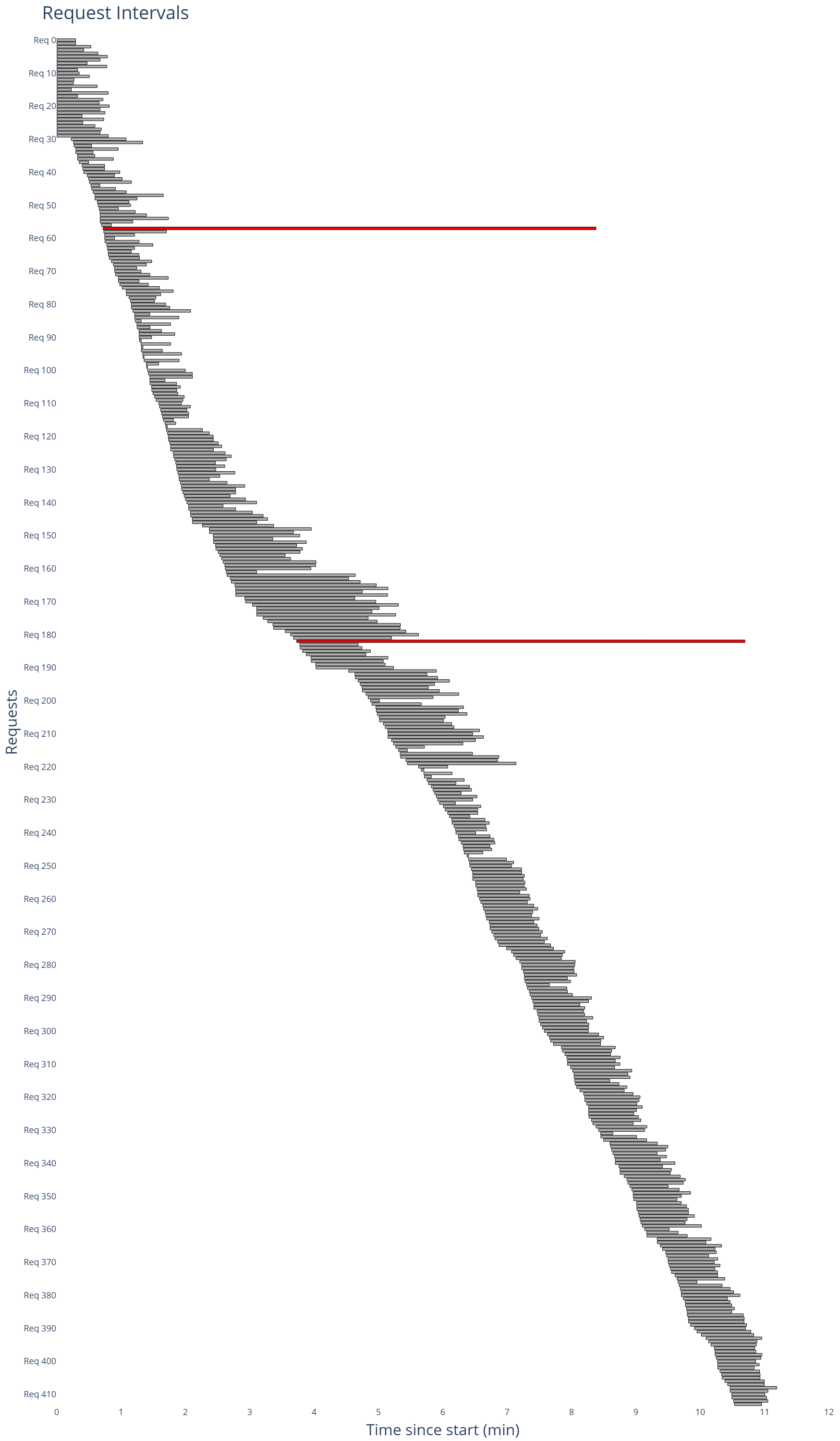}
  \captionof{figure}{Start and end time of each request (in submission order) for dataset 2. Each request is represented by a bar whose left edge marks start time and whose right edge marks end time. Degenerate requests are highlighted in red.}
  \label{fig:req_tempo_conjunto2}
\end{center}
\newpage}

\afterpage{%
\begin{center}
  \includegraphics[height=0.92\textheight]{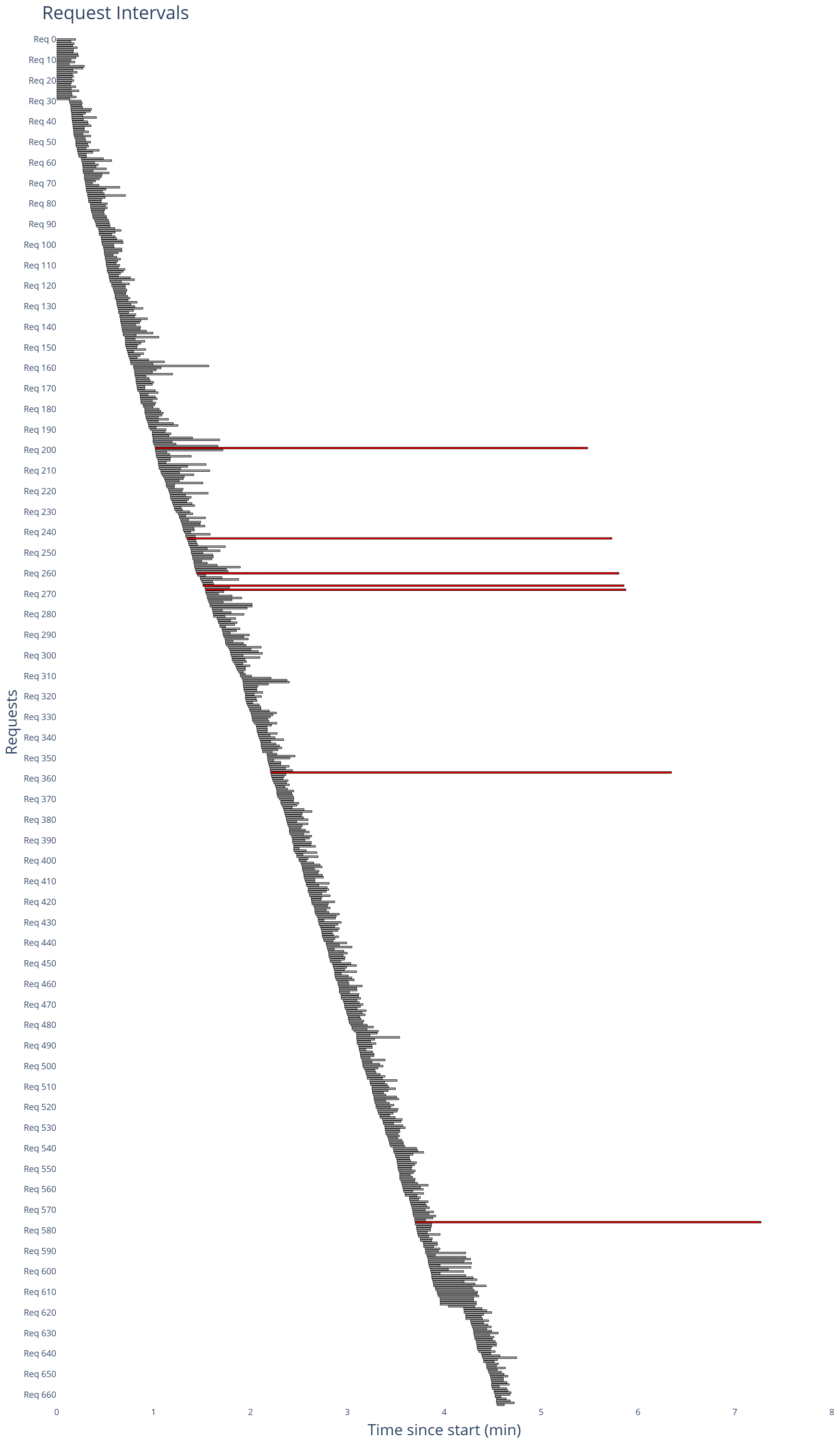}
  \captionof{figure}{Start and end time of each request (in submission order) for dataset 3. Each request is represented by a bar whose left edge marks start time and whose right edge marks end time. Degenerate requests are highlighted in red.}
  \label{fig:req_tempo_conjunto3}
\end{center}
\newpage}

As discussed in Section~\ref{sec:degeneracao-textual}, red bars last longer, keeping the model occupied for a disproportionate amount of time. In datasets 1 and 2, degeneration occurs in intermediate requests, but none of them finishes last. In dataset 3, by contrast, all non-degenerate requests are completed nearly three minutes before the overall end time and the remaining inference time is spent exclusively on degenerate requests.

The same scenario was simulated for the estimate of real impact of degenerate requests while degenerate requests were replaced with those whose duration equaled the mean duration of non-degenerate requests, yielding the results in the table below:

\begin{table}[htbp]
\centering
\caption{Comparison between observed inference time for each dataset and the time obtained by replacing degenerate requests with requests of average (non-degenerate) duration.}
\label{tab:tempo_degeneracao}

\begin{tabular}{
    p{2cm}
    p{2.6cm}
    p{3.7cm}
    p{2cm}
    p{3.5cm}
}
\toprule
\shortstack{\textbf{Dataset}} &
\shortstack{\textbf{Observed time}} &
\shortstack{\textbf{No-degeneration time}} &
\shortstack{\textbf{$\Delta$ time}} &
\shortstack{\textbf{Relative reduction}} \\
\midrule
1 & 21.7 min & 19.5 min & 2.2 min & 10.14 \% \\
2 & 11.2 min & 10.8 min & 0.4 min &  3.57 \% \\
3 &  7.3 min &  4.2 min & 3.1 min & 42.47 \% \\
\bottomrule
\end{tabular}

\end{table}

According to the table, even ignoring other negative effects of text degeneration on inference (e.g., throughput reduction and need for retries), replacing degenerate requests with average-duration ones can lead to an above 42\% reduction in wall-clock machine time, hence, a proportional reduction in machine cost.

Besides the simulation, the way a degenerate request affects the other requests in the dataset was analyzed. The duration distribution of non-degenerate requests was measured during periods when at least one degenerate request was active versus periods when none was. Since the excessive memory occupation induced by degeneration occurs precisely when the generated token sequence becomes longer than usual, a degeneration was considered ``in progress'' only after the degenerate request duration had exceeded the 0.99 percentile of the non-degenerate duration distribution. 

From that point onwards, healthy requests are affected by the abnormal GPU-memory occupation caused by the degenerate sequence. Figure~\ref{fig:boxplots_degeneracao} shows the duration distribution of healthy requests periods when at least one degenerate request in progress versus periods with none.

\afterpage{%
\clearpage
\begin{figure}[htbp]
  \centering
  \includegraphics[height=0.9\textheight]{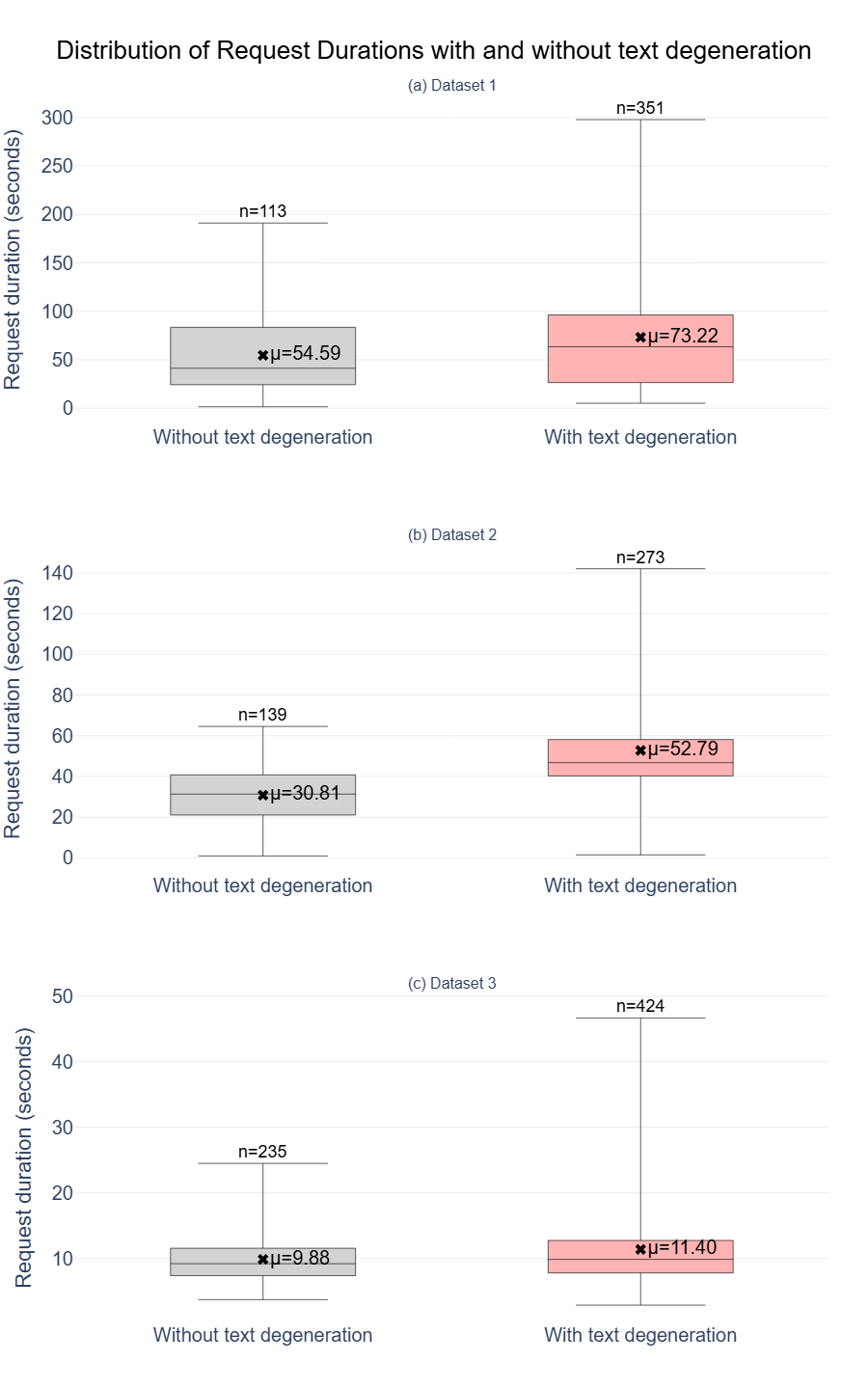}
  \caption{Distribution of healthy-request durations for the three datasets, contrasting periods with at least one degenerate request running in parallel versus periods with no degenerate request running. The mean of each distribution is marked in black with an ``x'' inside each box, with its value next to it.}
  \label{fig:boxplots_degeneracao}
\end{figure}
\clearpage
}

The plots show that, across all three datasets, the duration distribution of healthy requests observed while at least one degenerate request was in progress exhibits a higher mean, median, and spread than when no degenerate request was running, indicating healthy-request execution time is negatively affected by degenerate requests.

In more detail, for the three datasets, the mean duration of healthy requests running in parallel with at least one active degenerate request is at least 15\% higher than when no degenerate request is running, reaching more than 71\% in dataset 2. This highlights the negative impact a single degenerate request can have on other requests, increasing total inference time, hence, machine cost.

\subsection{Model Output Format}{Model Output Format}
\label{apx:output-format}

\begin{tcolorbox}[colback=white, colframe=black, boxrule=0.4pt]
\begin{verbatim}
{
    "header": "<header content>",
    "margin": "<margin content>",
    "footer": "<footer content>",
    "text":   "<main content>"
}
\end{verbatim}
\end{tcolorbox}
Structured output format used during dataset annotation. Fields absent from the original document are returned with null value \texttt{null}.

\begin{figure}[h]
    \centering
    \fbox{\includegraphics[width=0.85\linewidth]{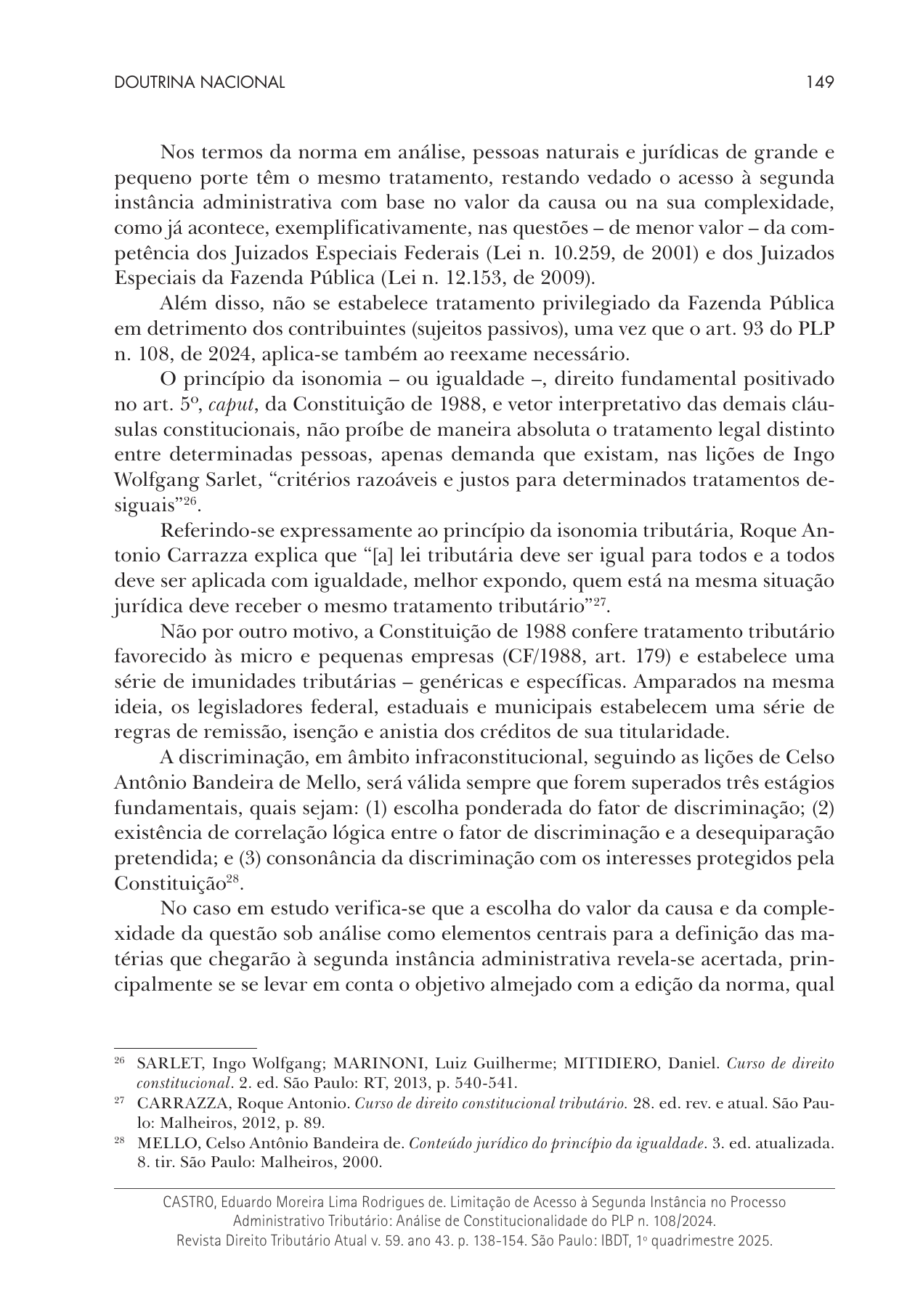}}
    \caption{Example of document used to illustrate the structured output format.}
    \label{fig:json_example}
\end{figure}

Figure~\ref{fig:json_example} shows an example page for extraction, illustrating a typical case of a document with well-defined peripheral structural metadata.

\clearpage
\begin{tcolorbox}[
  listing only,
  breakable,
  colback=white,
  colframe=black,
  boxrule=0.4pt,
  left=2mm,
  right=2mm,
  top=1mm,
  bottom=1mm,
  listing options={
      basicstyle=\ttfamily\small,
      breaklines=true,
      breakatwhitespace=true,
      columns=fullflexible,
      keepspaces=true,
      extendedchars=true,
      inputencoding=utf8
    }
]
\begin{lstlisting}
{
  "text": "Nos termos da norma em análise, pessoas naturais e jurídicas de grande e pequeno porte têm o mesmo tratamento, restando vedado o acesso à segunda instância administrativa com base no valor da causa ou na sua complexidade, como já acontece, exemplificativamente, nas questões de menor valor da competência dos Juizados Especiais Federais (Lei n. 10.259, de 2001) e dos Juizados Especiais da Fazenda Pública (Lei n. 12.153, de 2009).\nAlém disso, não se estabelece tratamento privilegiado da Fazenda Pública em detrimento dos contribuintes (sujeitos passivos), uma vez que o art. 93 do PLP n. 108, de 2024, aplica-se também ao reexame necessário.\nO princípio da isonomia ou igualdade, direito fundamental positivado no art. 5, caput, da Constituição de 1988, e vetor interpretativo das demais cláusulas constitucionais, não proíbe de maneira absoluta o tratamento legal distinto entre determinadas pessoas, apenas demanda que existam, nas lições de Ingo Wolfgang Sarlet, "critérios razoáveis e justos para determinados tratamentos desiguais"^26.\nReferindo-se expressamente ao princípio da isonomia tributária, Roque Antonio Carrazza explica que "[a] lei tributária deve ser igual para todos e a todos deve ser aplicada com igualdade, melhor expondo, quem está na mesma situação jurídica deve receber o mesmo tratamento tributário"^27.\nNão por outro motivo, a Constituição de 1988 confere tratamento tributário favorecido às micro e pequenas empresas (CF/1988, art. 179) e estabelece uma série de imunidades tributárias genéricas e específicas. Amparados na mesma ideia, os legisladores federal, estaduais e municipais estabelecem uma série de regras de remissão, isenção e anistia dos créditos de sua titularidade.\nA discriminação, em âmbito infraconstitucional, seguindo as lições de Celso Antônio Bandeira de Mello, será válida sempre que forem superados três estágios fundamentais, quais sejam: (1) escolha ponderada do fator de discriminação; (2) existência de correlação lógica entre o fator de discriminação e a desequiparação pretendida; e (3) consonância da discriminação com os interesses protegidos pela Constituição^28.\nNo caso em estudo verifica-se que a escolha do valor da causa e da complexidade da questão sob análise como elementos centrais para a definição das matérias que chegarão à segunda instância administrativa revela-se acertada, principalmente se se levar em conta o objetivo almejado com a edição da norma, qual\n^26 SARLET, Ingo Wolfgang; MARINONI, Luiz Guilherme; MITIDIERO, Daniel. Curso de direito constitucional. 2. ed. São Paulo: RT, 2013, p. 540-541.\n^27 CARRAZZA, Roque Antonio. Curso de direito constitucional tributário. 28. ed. rev. e atual. São Paulo: Malheiros, 2012, p. 89.\n^28 MELLO, Celso Antônio Bandeira de. Conteúdo jurídico do princípio da igualdade. 3. ed. atualizada. 8. tir. São Paulo: Malheiros, 2000.",
  "header": "DOUTRINA NACIONAL 149",
  "margin": null,
  "footer": "CASTRO, Eduardo Moreira Lima Rodrigues de. Limitação de Acesso à Segunda Instância no Processo Administrativo Tributário: Análise de Constitucionalidade do PLP n. 108/2024.\nRevista Direito Tributário Atual v. 59. ano 43. p. 138-154. São Paulo: IBDT, 1o quadrimestre 2025."
}
\end{lstlisting}
\end{tcolorbox}
Transcription generated by DharmaOCR Lite model, with a clear separation in JSON formatting.
\label{lst:apx-json-rosa}

Figure~\ref{fig:json_example_handwritten} presents a more demanding case, where handwritten text poses additional challenges for extraction.

\begin{figure}[h]
    \centering
    \fbox{\includegraphics[width=0.95\linewidth]{json_example_handwritten.png}}
    \caption{Example document used to illustrate structured output format for handwritten document.}
    \label{fig:json_example_handwritten}
\end{figure}

\clearpage
\begin{tcolorbox}[
  listing only,
  breakable,
  colback=white,
  colframe=black,
  boxrule=0.4pt,
  left=2mm,
  right=2mm,
  top=1mm,
  bottom=1mm,
  listing options={
      basicstyle=\ttfamily\small,
      breaklines=true,
      breakatwhitespace=true,
      columns=fullflexible,
      keepspaces=true,
      extendedchars=true,
      inputencoding=utf8
    }
]
\begin{lstlisting}
{
  "header": null,
  "text": "O filme Coringa retrata a história de Arthur, um homem solitário com diversos traumas\nde infância, que tenta lidar com a vida e com sua doença mental. No filme, Arthur salta\nem seu diário o fardo de ter uma doença psicológica e a pressão imposta pela sociedade para\nque ele a esconda. Fora da ficção, o estigma associado às doenças mentais na sociedade\nbrasileira não é tão disfarçado e decorre, principalmente, do tabu acerca de doenças psicológicas\ne da ausência de ações do Estado.\n\nEm primeira análise, observa-se que o preconceito enraizado dificulta o diagnóstico de doenças\nmentais. Nesse sentido, o psicanalista Sigmund Freud definiu como \"tabu\" certos temas con-\nsiderados sagrados, inquietantes e que não devem ser discutidos abertamente. Sob esta óptica, o\nestigma associado às doenças mentais advém de um tabu que, ao longo do tempo, reprimiu o\ndiálogo e o esclarecimento sobre o tema, enraizando o preconceito com doenças mentais na so-\nciedade brasileira. Dessa forma, é crível que tal ambiente social inibe a possibilidade de diagno-\nticos e tratamentos adequados, conduzindo um espaço onde milhões de brasileiros lidam soze-\nnhos com doenças psicológicas.\n\nOutrossim, destaca-se que o dever do Estado é minimizar os impasses para garantir a qua-\nlidade de vida dos cidadãos. Consoante o químico H. Louis Le Chatelier, \"quando um sistema em\nequilíbrio recebe algum tipo de perturbação externa, ele se deslocará no sentido de minimizar es-\nsa perturbação e retornar ao estado de equilíbrio\". Sob este viés, apesar de ser um princípio quí-\nmico, ele também pode ser aplicado ao meio social, uma vez que o dever do Estado é tomar\nmedidas para que o estigma associado às doenças mentais seja atenuado. Desse modo, é impres-\ncindível que os órgãos estatais solucionem o óbice e efetivem a qualidade de vida dos cidadãos.\n\nPortanto, a respeito do estigma associado às doenças mentais na sociedade brasileira é mister\nque o Estado tome medidas para solução do entrave. Para tal, compete ao Ministério da Edu-\ncação elaborar eventos abertos para a comunidade, em escolas públicas e privadas, em que\npor meio de palestras, psicólogos e professores discursarão a respeito da saúde e de doen-\nças mentais. Além disso, visto que atividades culturais possuem imenso poder transfor-\nmador, tais eventos deverão proporcionar para os crianças brincadeiras lúdicas e edu-\ncacionais, visando-as a identificar doenças psicológicas e, aos pais, como procurar\ntratamento, para que assim a realidade social no Brasil seja diferente da que é\nretratada no filme Coringa.",
  "footer": null,
  "margin": null
}
\end{lstlisting}
\end{tcolorbox}
Transcription generated by the DharmaOCR Lite model, showcasing a more complex task of understanding handwritten text.
\label{lst:apx-json-handwritten}

\clearpage
\subsection{Evaluation Criteria for DPO Annotation}
\label{apx:criterios-avaliacao}

The four criteria for the evaluation of generated responses are described below. Each criterion was presented to the evaluator model as an objective question, as specified by the prompts in Table~\ref{tab:criterios}.

\begin{table}[htbp]
\centering
\caption{Prompts for the evaluation criteria used during labeling.}
\label{tab:criterios}
\begin{tabular}{p{2cm} p{11cm}}
\toprule
\textbf{\shortstack{Criterion}} & \textbf{Prompt} \\
\midrule
\shortstack{Completeness} &
Is the extraction complete?, i.e., did the model detect and extract all text present in the image without omitting characters, words, lines, or text blocks? Does the extraction cover all textual elements visible in the original image? \\
\midrule
\shortstack{Precision} &
Is the extraction precise?, i.e., does the extracted text exactly match the original text with no typos, character substitutions, letter inversions, or hallucinated words that are not present in the image? \\
\midrule
\shortstack{Formatting} &
Does the extraction preserve the appropriate formatting?, i.e., did the model maintain the original text's visual structure, respecting line breaks, paragraphs, spacing, alignment, and the overall layout of the content as it appears in the image? \\
\midrule
\shortstack{Structure \\ Adherence} &
Does the extraction adhere to the expected \texttt{json} structure? The expected structure is \texttt{\{"text": "extracted\_text", "header": "header\_contet", "margin": "margin\_contet", "footer": "footer\_contet"\}}. If \texttt{json} format is respected, does the extraction correctly assign values to the corresponding fields, assigning null values to fields absent from the image? \\
\bottomrule
\end{tabular}
\end{table}

\clearpage
\subsection{Filtering and Selection Policy for Preference Pairs}
\label{apx:filtragem-pares}

\begin{enumerate}
  \item The aggregated score was computed for each instance and each response as the arithmetic mean of the four criterion scores.
  
  \item All-vs-all pairing among the five responses (10 pairs per instance) was generated for each instance, yielding the 237\,260 candidate pairs previously reported.
  
  \item A multi-stage filtering policy, summarized in Table~\ref{tab:criterios-filtragem} and detailed below. In general terms, it pursues two complementary objectives, namely, (i) to ensure the included pairs provide an instructive signal for preference learning (maximize signal / reduce noise) and (ii) to avoid pairs that induce optimization conflicts or probability-shift effects, as identified in recent work on DPO and its variants.

\begin{itemize}
  \item \textbf{Filtering excessively difficult examples (Selective-DPO):} following the principle demonstrated by Gao et al.\ \cite{gao2025principled}, pairs in which both responses have low quality tend to introduce noise and harm alignment, since neither provides a reliable instructive target, yielding contradictory signals in the DPO gradient. Therefore, pairs in which both responses scored below 800/1000 were discarded - in such cases, the inclusion of the pair increases signal variance without providing a clear improvement target.

  \item \textbf{Disparity criterion for mid-quality regime:} when neither response in the pair exceeds 900 points, but at least one exceeds 800, a minimum disparity of 400 points between them is required. Such a motivation is twofold, i.e., (a) pairs with large disparity in this regime contain a strong relative signal, enabling the model to distinguish undesirable patterns even if the preferred response (\texttt{chosen}) is not perfect; and (b) pairs with small differences in this regime tend to produce reduced reward margins—a situation that can lead to optimization conflict \cite{xu2025beyond}, in which the updates for the preferred (\texttt{chosen}) and rejected (\texttt{rejected}) responses become similar, impairing learning.

  \item \textbf{Policy-guided threshold for high-quality cases:} when at least one response in the pair achieves a score of 900 or a higher one, the pair is considered potentially highly informative; nevertheless, a minimum difference of 200 points is required to avoid inclusion of ambiguous choices. Such a constraint is directly inspired in the analyses of Xu et al.\ \cite{xu2025beyond} on the effects of reward margins in DPO dynamics, i.e., very small margins tend to induce optimization conflict. Although Xu et al.\ also pointed out risks associated with excessively large margins, a maximum-margin filter was not applied in this study, considering high-disparity cases among high-quality pairs carry relevant instructive signal and the implicit regularization of DPO's $\beta$ is sufficient to attenuate potential gradient-saturation effects.

  \item \textbf{Chosen/rejected assignment and retention of degeneration cases:} once a pair as passed the aforementioned filters, the higher-scoring response is labeled as \texttt{chosen} and the other is \texttt{rejected}. Instances of text degeneration were deliberately preserved among the \texttt{rejected} examples, provided those responses did not appear as \texttt{chosen} in other pairs. Degeneration examples constitute behavioral patterns that should be mitigated and their explicit penalization reinforces that effect. From an optimization standpoint, the consistent presence of those low-scoring \texttt{rejected} responses creates a clear negative signal that pushes the policy to reducing undesirable continuations.
\end{itemize}
\end{enumerate}

\begin{table}[htbp]
\centering
\caption{Filtering and selection criteria for preference pairs, where $s_w$ and $s_l$ denote aggregated scores of \texttt{chosen} and \texttt{rejected} responses in the pair, respectively.}
\label{tab:criterios-filtragem}
\begin{tabular}{p{3.5cm} p{3.5cm} p{3.5cm} p{2cm}}
\toprule
\textbf{\shortstack{Condition}} &
\textbf{\shortstack{Requirement}} &
\textbf{\shortstack{Motivation}} &
\textbf{\shortstack{Action}} \\
\midrule
$s_w < 800$ and $s_l < 800$ &
— &
Unreliable signal &
Discard \\
\midrule $\max(s_w, s_l) \geq 800$ and $\max(s_w, s_l) < 900$) &
$s_w - s_l \geq 400$ &
Avoid small margins in mid-quality regime &
Include \\
\midrule
$\max(s_w, s_l) \geq 900$ &
$s_w - s_l \geq 200$ &
Informative pair with a guaranteed minimum margin &
Include \\
\bottomrule
\end{tabular}
\end{table}

\clearpage

\subsection{Cost Computation Details}
\label{apx:calculo-custos}

This subsection documents the way the cost per million processed pages were computed for each model.

\subsubsection{Local models (DharmaOCR and open-source baselines)}

According to the average time per page $T_{avg}$ measured during the benchmark and the hourly price $P_{hour}$ of the inference infrastructure used, the cost per million pages were estimated as

\begin{equation}
C_{1M} = 10^6\,T_{avg}\left(\frac{P_{hour}}{3600}\right)
\end{equation}

\subsubsection{Amazon Textract and Google Vision}

Pricing is fixed per thousand pages (except for volume-dependent discounts, which were ignored here) for Amazon Textract and Google Vision. Consequently, both were priced at USD 1.50 per thousand pages (as of March 2026) and $C_{1M}$ was computed as

\begin{equation}
    C_{1M} = 10^6 \left(\frac{1{,}50}{1000}\right)
\end{equation}

\subsubsection{Google Document AI}
Similarly to Google Vision, the pricing of Google Document AI Custom Extractor with generative AI is fixed per thousand pages, however, it is higher due to the use of generative AI. The price (as of March 2026) is USD 30.00 per thousand pages; therefore, $C_{1M}$ for this model, was computed as

\begin{equation}
    C_{1M} = 10^6 \left(\frac{30}{1000}\right)
\end{equation}

\subsubsection{Mistral OCR3}

Pricing is also fixed per thousand pages, at USD 2.00 per thousand pages (as of March 2026) for Mistral OCR 3. Therefore, $C_{1M}$ was computed as

\begin{equation}
    C_{1M} = 10^6 \left(\frac{2}{1000}\right)
\end{equation}

\subsubsection{GPT-4o}

\par
The observed average number of tokens per page (input and output) and the prices per million tokens (as of March 2026) were used for token-priced APIs for estimating an average per-page cost, which was then multiplied by one million to obtaining the cost per million pages. Denoting the average numbers of input and output tokens per page by $n_{in}$ and $n_{out}$ and the corresponding prices per million tokens by $p_{in}$ and $p_{out}$ one has

\begin{equation}
    C_{1M} = 10^6 \left(\frac{n_{in} \, p_{in}}{10^6} + \frac{n_{out} \, p_{out}}{10^6} \right)
\end{equation}

\clearpage
\raggedbottom
\subsection{Detailed OCR Benchmark Results for Quantized Models}

\label{apx:quant-benchmark-ocr}

In what follows are the results of experiments that quantized the best models obtained during training with use of AWQ at different precision levels. All quantizations used a calibration dataset composed of one thousand samples from the dataset used for model training.

\begin{table}[htbp]
\centering
\caption{Performance Analysis: AWQ-Quantized Models}
\label{tab:quantizacao}
\footnotesize
\setlength{\tabcolsep}{3pt}
\begin{tabular}{@{}
  >{\raggedright\arraybackslash}m{4.9cm}
  >{\raggedright\arraybackslash}m{2.4cm}
  >{\centering\arraybackslash}m{1.8cm}
  >{\centering\arraybackslash}m{2.1cm}
  >{\centering\arraybackslash}m{1.9cm}
@{}}
\toprule
\textbf{\shortstack{Model}} & \textbf{\shortstack{Technique}} & \textbf{Benchmark Score} & \textbf{\shortstack{Degeneration\\Rate (\%)}} & \textbf{\shortstack{Time per\\page (s)}} \\ \midrule

olmOCR-2-7B (SFT + DPO) & AWQ W8A8 & 0.926 & 0.60 & 2.122 \\
olmOCR-2-7B (SFT + DPO) & AWQ FP8 & 0.925 & 1.21 & 2.187 \\
olmOCR-2-7B (SFT + DPO) & AWQ W4A16 & 0.903 & 1.61 & 2.004 \\
\midrule
Nanonets-OCR2-3B (SFT + DPO) & AWQ FP8 & 0.912 & 0.20 & 1.463 \\
Nanonets-OCR2-3B (SFT + DPO) & AWQ W8A8 & 0.903 & 1.81 & 1.500 \\
Nanonets-OCR2-3B (SFT + DPO) & AWQ W4A16 & 0.831 & 11.09 & 1.821 \\ \bottomrule

\end{tabular}
\end{table}

\raggedbottom
\subsection{LLM-as-a-Judge Methodology with GPT-4o}
\label{apx:llm-as-judge}

Implemented an LLM-as-a-Judge approach using the GPT-4o model for the qualitative and comparative assessment of text-extraction results. The process consists of a pairwise comparison in which the evaluator simultaneously analyzes the original image and the outputs produced by two distinct models.

\subsubsection{Bias Mitigation}

The following strategies were adopted to ensure integrity of the results and mitigate common biases in language models:

\begin{itemize}
    \item \textbf{Anonymization}: Models are identified only by random IDs (e.g., 5DA and 18C), preventing the judge from inferring with the architecture or the model name.

    \item \textbf{Double execution}: Each comparison was performed twice towards reducing the effects of LLM stochasticity and a response was considered better only if the judge remained consistent across both rounds.

    \item \textbf{Order shuffling}: Each pair of responses was evaluated twice in different orders, i.e., once as Model X vs. Model Y and once as Model Y vs. Model X, for the countering of order-related bias.
\end{itemize}

\subsubsection{Evaluation Criteria}

The evaluation was divided into six key dimensions, described in Table~\ref{tab:criterios_judge}.

\begin{table}[!ht]
\centering
\caption{Technical Evaluation Criteria (GPT-4o Judge)}
\label{tab:criterios_judge}
\footnotesize
\begin{tabular}{ll p{9cm}}
\toprule
\textbf{ID} & \textbf{\shortstack{Criterion}} & \textbf{\shortstack{Guiding Question}} \\ \midrule
C1 & Comprehensibility & Which text enables information to be extracted more easily? \\
C2 & Precision & Which text exhibits the lowest rate of extraction errors? \\
C3 & Completeness & Which model extracted the largest amount of relevant information? \\
C4 & Fidelity & Which text is more faithful, avoiding hallucinated words that are not present? \\
C5 & General Adequacy & Which model is more robust for deployment in production systems? \\
C6 & Formatting & Which model better preserves the structure? \\ \bottomrule
\end{tabular}
\end{table}

The criteria were included in the judge LLM prompt so that it could determine the best response (or if there was a tie) for each criterion.

\clearpage
\raggedbottom
\subsection{Additional Examples of Text Degeneration}
\label{apx:examples-degeneration}
\noindent\begin{tcolorbox}[breakable, colback=gray!10, colframe=gray!50, boxrule=0.4pt, before skip=0pt, after skip=0pt]
\begin{lstlisting}[basicstyle=\ttfamily\scriptsize, breaklines=true]
**BREAK-POINT TESTS**

All the organisms were tested against all the drugs on five occasions and the number appearing resistant (R), moderate (M) or sensitive (S) in all five tests in
4/5 tests and in 3/5 tests are given in Table 2. It will be seen that a very large proportion of strains varied between S and M to CLOR and CTIN. and results
with CFOX and CROX were little better. This variation is almost certainly the result of the proximity of the lower break-point used (4 mg/l) to the MIC of many
of the strains (see Table 1). In contrast, the results with CMAN, CZOL and CTAX were much more consistant because the MICs were considerably lower. The MIC of
the beta-lactamase-producing strains was 4-8 mg/l; two strains were consistently S to CLOR. two varied between S and M and 10 were always M.

There is no evidence that the results were affected by either of the media used; both appeared satisfactory and the swarming of Proteus spp was adequately
controlled.

**EFFECT OF INOCULUM**

Plates containing 16 and 4 mg/l of CLOR, CTIN, CLEX, CFOX and CROX were inoculated with overnight broth cultures diluted 1/10 and 1/10(X); 102 strains were
tested. With CLOR 13 strains changed from S to M and 10 from M to R with the heavier inoculum, but these included only three of the beta-lactamase-producing E.
coli, the remainder being unaffected. With CTIN six strains changed from S to M and 10 from M to R. Six strains changed from S to

| Cephalosporin | No of strains inhibited by eight cephalosporins (mg/ml) |
|---------------|-----------------------------------------------------|
|               | Resistant | Moderate | Sensitive |
|               | >256 | 256 | 128 | 64 | 32 | 16 | 8 | 4 | 2 | 1 | 0.5 |
| Cephaloridine (CLOR) | 16 | 9 | 5 | 6 | 3 | 2 | 1 | 2 | 50 | 24 |
| Cephalothin (CTIN) | 15 | 8 | 6 | 5 | 1 | 37 | 31 | 15 | 2 |
| Cephalexin (CLEX) | 14 | 1 | 9 | 6 | 2 | 13 | 45 | 35 | 2 |
| Cefoxitin (CFOX) | 2 | 1 | 6 | 3 | 5 | 4 | 18 | 27 | 12 |
| Cefuroxime (CROX) | 6 | 3 | 3 | 5 | 4 | 14 | 8 | 11 | 8 |
| Cefamandole (CMAN) | 4 | 2 | 6 | 5 | 4 | 8 | 8 | 9 | 71 |
| Cefazolin (CZOL) | 12 | 7 | 5 | 3 | 1 | 8 | 11 | 5 |
| Cefotaxime (CTAX) | 2 | 1 | 2 | 1 | 2 | 1 | 2 | 1 | 2 | 1 | 2 | 1 | 2 | 1 | 2 | 1 | 2 | 1 | 2 | 1 | 2 | 1 | 2 | 1 | 2 | 1 | 2 | 1 | 2 | 1 | 2 | 1 | 2 | 1 | 2 |
| 1 | 2 | 1 | 2 | 1 | 2 | 1 | 2 | 1 | 2 | 1 | 2 | 1 | 2 | 1 | 2 | 1 | 2 | 1 | 2 | 1 | 2 | 1 | 2 | 1 | 2 | 1 | 2 | 1 | 2 | 1 | 2 | 1 | 2 | 1 | 2 | 1 | 2 | 1 | 2 |
| 1 | 2 | 1 | 2 | 1 | 2 | 1 | 2 | 1 | 2 | 1 | 2 | 1 | 2 | 1 | 2 | 1 | 2 | 1 | 2 | 1 | 2 | 1 | 2 | 1 | 2 | 1 | 2 | 1 | 2 | 1 | 2 | 1 | 2 | 1 | 2 | 1 | 2 | 1 | 2 |
| 1 | 2 | 1 | 2 | 1 | 2 | 1 | 2 | 1 | 2 | 1 | 2 | 1 | 2 | 1 | 2 | 1 | 2 | 1 | 2 | 1 | 2 | 1 | 2 | 1 | 2 | 1 | 2 | 1 | 2 | 1 | 2 | 1 | 2 | 1 | 2 | 1 | 2 | 1 | 2 |
| 1 | 2 | 1 | 2 | 1 | 2 | 1 | 2 | 1 | 2 | 1 | 2 | 1 | 2 | 1 | 2 | 1 | 2 | 1 | 2 | 1 | 2 | 1 | 2 | 1 | 2 | 1 | 2 | 1 | 2 | 1 | 2 | 1 | 2 | 1 | 2 | 1 | 2 | 1 | 2 |
| 1 | 2 | 1 | 2 | 1 | 2 | 1 | 2 | 1 | 2 | 1 | 2 | 1 | 2 | 1 | 2 | 1 | 2 | 1 | 2 | 1 | 2 | 1 | 2 | 1 | 2 | 1 | 2 | 1 | 2 | 1 | 2 | 1 | 2 | 1 | 2 | 1 | 2 | 1 | 2 |
| 1 | 2 | 1 | 2 | 1 | 2 | 1 | 2 | 1 | 2 | 1 | 2 | 1 | 2 | 1 | 2 | 1 | 2 | 1 | 2 | 1 | 2 | 1 | 2 | 1 | 2 | 1 | 2 | 1 | 2 | 1 | 2 | 1 | 2 | 1 | 2 | 1 | 2 | 1 | 2 |
| 1 | 2 | 1 | 2 | 1 | 2 | 1 | 2 | 1 | 2 | 1 | 2 | 1 | 2 | 1 | 2 | 1 | 2 | 1 | 2 | 1 | 2 | 1 | 2 | 1 | 2 | 1 | 2 | 1 | 2 | 1 | 2 | 1 | 2 | 1 | 2 | 1 | 2 | 1 | 2 |
| 1 | 2 | 1 | 2 | 1 | 2 | 1 | 2 | 1 | 2 | 1 | 2 | 1 | 2 | 1 | 2 | 1 | 2 | 1 | 2 | 1 | 2 | 1 | 2 | 1 | 2 | 1 | 2 | 1 | 2 | 1 | 2 | 1 | 2 | 1 | 2 | 1 | 2 | 1 | 2 |
| 1 | 2 | 1 | 2 | 1 | 2 | 1 | 2 | 1 | 2 | 1 | 2 | 1 | 2 | 1 | 2 | 1 | 2 | 1 | 2 | 1 | 2 | 1 | 2 | 1 | 2 | 1 | 2 | 1 | 2 | 1 | 2 | 1 | 2 | 1 | 2 | 1 | 2 | 1 | 2 |
| 1 | 2 | 1 | 2 | 1 | 2 | 1 | 2 | 1 | 2 | 1 | 2 | 1 | 2 | 1 | 2 | 1 | 2 | 1 | 2 | 1 | 2 | 1 | 2 | 1 | 2 | 1 | 2 | 1 | 2 | 1 | 2 | 1 | 2 | 1 | 2 | 1 | 2 | 1 | 2 |
...
\end{lstlisting}
\end{tcolorbox}

\begin{tcolorbox}[colback=gray!10, colframe=gray!50, boxrule=0.4pt]
\begin{lstlisting}[basicstyle=\ttfamily\scriptsize, breaklines=true]
Resultado Líquido 9.867 23.297 (50)% 
Ebitda 12.818 17.476 (25)% 
Liquidez (ativos menos passivos) 7.340 7.288 0.8% 
Resultado por Ação 1.070 1.357 (22)% 
Risco de Reputação 27 5 475% 
Resultado Não Operacional 1.282 543 136% 
Resultado Líquido 9.867 23.297 (50)% 
Ebitda 12.818 17.476 (25)% 
Liquidez (ativos menos passivos) 7.340 7.288 0.8% 
Resultado por Ação 1.070 1.357 (22)% 
Risco de Reputação 27 5 475% 
Resultado Não Operacional 1.282 543 136% 
Resultado Líquido 9.867 23.297 (50)% 
Ebitda 12.818 17.476 (25)% 
Liquidez (ativos menos passivos) 7.340 7.288 0.8% 
Resultado por Ação 1.070 1.357 (22)%
...
\end{lstlisting}
\end{tcolorbox}

\begin{tcolorbox}[colback=gray!10, colframe=gray!50, boxrule=0.4pt]
\begin{lstlisting}[basicstyle=\ttfamily\scriptsize, breaklines=false]
Sumário:

PREFÁCIO .................................................................. 09  
APRESENTAÇÃO .............................................................. 45  
CAPÍTULO I - HISTÓRICO .................................................... 67 
CAPITULOS II FUNÇÕES ...................................................... 98
CAPÍTULO III - CÁLCULO DIFFERENCIAL ...............................................................
...................................................................................................
...................................................................................................
...................................................................................................
...................................................................................................
...................................................................................................
...............................................................................
...
\end{lstlisting}
\end{tcolorbox}

\begin{tcolorbox}[colback=gray!10, colframe=gray!50, boxrule=0.4pt]
\begin{lstlisting}[basicstyle=\ttfamily\scriptsize, breaklines=false]
Quarta-feira, 24 de novembro de 2004

Poder Judiciário - Caderno 3 - Parte II

---
São Paulo, 74 (217) - 229
---
São Paulo, 74 (217) - 229
---
São Paulo, 74 (217) - 229
---
São Paulo, 74 (217) - 229
---
São Paulo, 74 (217) - 229
...
\end{lstlisting}
\end{tcolorbox}

\end{document}